\pdfoutput=1

\documentclass[11pt]{article}

\usepackage[]{acl}

\usepackage{times}
\usepackage{latexsym}
\usepackage{amsmath}
\usepackage{hyperref}
\usepackage{cleveref}
\usepackage{subcaption}
\usepackage{supertabular}
\usepackage{booktabs}
\usepackage{tabularx}
\usepackage{multirow}
\usepackage{multicol}
\usepackage{makecell}
\usepackage{soul}

\DeclareSymbolFont{extraup}{U}{zavm}{m}{n}
\DeclareMathSymbol{\vardiamond}{\mathalpha}{extraup}{87}

\newcommand\blfootnote[1]{%
  \begingroup
  \renewcommand\thefootnote{}\footnote{#1}%
  \addtocounter{footnote}{-1}%
  \endgroup
}

\usepackage[T1]{fontenc}

\usepackage[utf8]{inputenc}
\usepackage{CJKutf8}
\newcommand{\chin}[1]{\begin{CJK*}{UTF8}{gkai}#1\end{CJK*}}
\newcommand{\abchin}[1]{\begin{CJK*}{UTF8}{bkai}#1\end{CJK*}}

\usepackage{fix-cm}    

\makeatletter
\newcommand\HUGE{\@setfontsize\Huge{50}{60}}
\makeatother
\usepackage{graphicx}
\usepackage[normalem]{ulem}
\useunder{\uline}{\ul}{}

\usepackage{microtype}

\title{RoleEval: A Bilingual Role Evaluation Benchmark for Large Language Models}

\author{Tianhao Shen$^1$, Sun Li$^2$, Quan Tu$^3$, Deyi Xiong$^{1*}$ \\
$^1$ College of Intelligence and Computing, Tianjin University, Tianjin, China \\
$^2$ China Academy of Information and Communications Technology, Beijing, China \\
$^3$ Gaoling School of Artificial Intelligence, Renmin University of China \\
\texttt{\{thshen,dyxiong\}@tju.edu.cn} \\
\texttt{lisun@caict.ac.cn} \\
\texttt{quantu@ruc.edu.cn}
}

\begin{document}
\maketitle
\begin{abstract}
The rapid evolution of large language models necessitates effective benchmarks for evaluating their role knowledge, which is essential for establishing connections with the real world and providing more immersive interactions. This paper introduces RoleEval, a bilingual benchmark designed to assess the memorization, utilization, and reasoning capabilities of role knowledge. RoleEval comprises \textit{RoleEval-Global} (including internationally recognized characters) and \textit{RoleEval-Chinese} (including characters popular in China), with 6,000 Chinese-English parallel multiple-choice questions focusing on 300 influential people and fictional characters drawn from a variety of domains including celebrities, anime, comics, movies, TV series, games, and fictions. These questions cover basic knowledge and multi-hop reasoning abilities, aiming to systematically probe various aspects such as personal information, relationships, abilities, and experiences of the characters. To maintain high standards, we perform a hybrid quality check process combining both automatic and human verification, ensuring that the questions are diverse, challenging, and discriminative.

Our extensive evaluations with RoleEval across various open-source and proprietary large language models, under both the zero- and few-shot settings, reveal insightful findings. Notably, while GPT-4 outperforms other models on \textit{RoleEval-Global}, Chinese large language models excel on \textit{RoleEval-Chinese}, highlighting significant knowledge distribution differences. We expect that RoleEval would highlight the significance of assessing role knowledge for large language models across various languages and cultural settings.\blfootnote{* Corresponding author.}\footnote{Our dataset is available at \url{https://github.com/Magnetic2014/RoleEval}.}

\end{abstract}

\section{Introduction}

Recent years have witnessed the huge success of large language models (LLMs), and agents based on these models present immense potential for reshaping our engagement with machines with their expansive knowledge and remarkable predictive capabilities \citep{wangSurveyLargeLanguage2023, xiRisePotentialLarge2023}. The cornerstone of this transformation lies in the development of LLM agents with a keen perception of the real world, offering users a more immersive experience than ever before and laying a solid foundation for the emergence of applications such as Character AI\footnote{\url{https://beta.character.ai}} and AI Dungeon.\footnote{\url{https://aidungeon.com}} This necessitates the role-playing capabilities of these models, which aim to establish connections with real-world people or characters created by LLMs.

Previous studies show that much of LLM knowledge is acquired during the pretraining phase \citep{zhouLIMALessMore2023a, linUnlockingSpellBase2023}. A comprehensive pretraining allows for better role-playing capabilities by covering a wide range of knowledge. This is crucial not just for representing existing characters but also for creating new, believable personas without prior history. Just as actors need a solid understanding of real-world facts to convincingly portray fictional characters, learning from real-world personas enables a model to authentically and consistently represent both real and imaginary characters.

However, there is a shortage of systematic evaluation of role knowledge for these LLMs. Traditional persona-based evaluation benchmarks, such as PersonaChat \citep{zhangPersonalizingDialogueAgents2018} and PersonalDialog \citep{zhengPersonalizedDialogueGeneration2020}, often rely on artificially constructed personas or occupations abstracted from a group of people, which lack the complexity and real-world connection of genuine personas. Thus it is hard for using such conventional benchmarks to assess the models' capability in handling intricate character knowledge. Recent role-playing benchmarks like RoleBench \citep{wangRoleLLMBenchmarkingEliciting2023} evaluate consistency in language style and knowledge extracted from scripts. However, the knowledge extracted from scripts is fragmented and lacks systematicness, making it challenging to accurately and comprehensively assess the breadth of knowledge captured by LLMs.

\begin{figure}
    \centering
    \includegraphics[width=\linewidth]{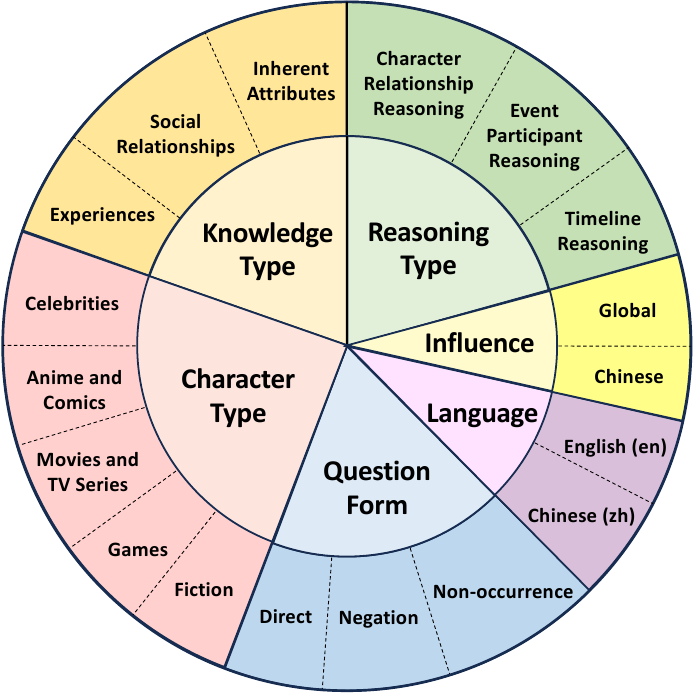}
    \caption{RoleEval includes different types of characters, knowledge, reasoning, question forms, influence, and languages.}
    \label{fig:overall_pie_chart}
\end{figure}

In light of these issues, we introduce RoleEval, a bilingual benchmark designed to assess the capturing and understanding of role knowledge related to both real-world and fictional characters for LLMs. RoleEval is comprised of two sub-benchmarks: \textit{RoleEval-Global} and \textit{RoleEval-Chinese}, with 300 characters and 6,000 Chinese-English parallel questions in total. Characters in RoleEval are from five categories: 1) celebrities, 2) anime and comics, 3) movies and TV series, 4) games, and 5) fictions. Each character is associated with 20 questions (17 focusing on basic knowledge and 3 on multi-hop reasoning), which evaluate the knowledge and capacity of LLMs in understanding personal information, relationships, abilities, and experiences, as illustrated in Figure \ref{fig:overall_pie_chart}. These questions are initially crafted in Chinese, then translated into English to construct bilingual parallel data.

Specifically, \textit{RoleEval-Global} is structured around 200 internationally recognized characters with 4,000 questions, which is suitable for evaluating both English and Chinese LLMs. \textit{RoleEval-Chinese} is targeting another set of 100 influential characters in China with 2,000 questions, following the same question construction framework as \textit{RoleEval-Global}, to specifically evaluate Chinese LLMs. The benchmark incorporates both automatic and manual quality controls to maintain high-quality and challenging questions.

To the best of our knowledge, RoleEval is the first benchmark to systematically evaluate the role knowledge for LLMs. It aims to benchmark the current state of LLMs in capturing, understanding, utilizing, and reasoning over knowledge of a wide range of both public figures and fictional characters. By delving into the depths of how LLMs perceive and portray an extensive array of characters, we aim to uncover new avenues for their application, making them not only repositories of information but also active, context-aware participants in our digital narratives.

In a nutshell, our contributions are as follows:
\begin{itemize}
    \item We propose RoleEval, a bilingual role evaluation benchmark with 6,000 Chinese-English parallel questions covering 300 diverse characters, to systematically examine the ability of capturing, understanding, and reasoning over role knowledge for LLMs, which is an important prerequisite for successful role-playing.
    \item To ensure quality and boost efficiency, we propose a hybrid quality check process with the combination of both automatic and manual verification to ensure appropriate difficulty and discrimination ability control for questions.
    \item We conduct extensive evaluations using RoleEval on a variety of large language models under both zero- and few-shot settings, encompassing models with varying parameter sizes and those mainly designed for English and Chinese, as well as both open-source and closed-source proprietary models.
\end{itemize}

\section{Related Work}

Role-playing agents \citep{shusterAmMeYou2021, liChatHaruhiRevivingAnime2023, shaoCharacterLLMTrainableAgent2023, wangRoleLLMBenchmarkingEliciting2023}, which can be even traced back to ELIZA \citep{weizenbaumELIZAComputerProgram1966}, the first automated dialogue agent that conducts psychological consulting, have recently attracted growing interest in both industry and academia. However, their evaluation is predominantly conducted on models after supervised fine-tuning. This approach does not incorporate direct feedback from pretrained base models, which can offer critical insights into their intrinsic role-playing capabilities and limitations. On the other hand, existing evaluations largely rely on outputs from the ChatGPT \citep{ouyangTrainingLanguageModels2022} or humans. However, ChatGPT is not an infallible evaluator, and human evaluation lacks reproducibility. This leads to a lack of objective, accurate, and systematic knowledge assessments. In this case, our benchmark can serve as a robust standard for evaluating current role-playing agents, assessing whether the models possess sufficient role knowledge.

Previous research in role-playing evaluation has largely focused on abstract personas \citep{zhangPersonalizingDialogueAgents2018, zhengPersonalizedDialogueGeneration2020, xuLongTimeNo2022a, weiMultiPartyChatConversational2023, ahnMPCHATMultimodalPersonaGrounded2023, zhouCharacterGLMCustomizingChinese2023} or specific professions like psychology consultants, chemists, or software engineers \citep{liuEmotionalSupportDialog2021, wangSurveyLargeLanguage2023}. However, these methods often oversimplify real-world personas, failing to capture the complex nature of real-world human personalities and behaviors in role-playing scenarios. Despite the value of these approaches, they fall short in exploring individual character knowledge, which is critical for authentic role-playing experiences. Therefore, there is a clear need for more sophisticated and realistic persona models in role-playing research. There are also some closely related works for character-based evaluation, intending to inspect the ability to mimic real-world characters \citep{shaoCharacterLLMTrainableAgent2023, wangRoleLLMBenchmarkingEliciting2023, zhouCharacterGLMCustomizingChinese2023}. However, they do not have a detailed and systematic framework to evaluate role knowledge, leading to fragmented and incomplete knowledge assessment. Moreover, their evaluation relies on humans or other powerful large language models such as ChatGPT. However, as we have stated above, this kind of evaluation suffers from the reproducibility and accuracy of judges. In contrast, RoleEval examines the knowledge required for role play in a detailed, objective and systematic approach for LLMs.

\section{RoleEval}
We develop RoleEval to assess how well role knowledge is captured, utilized, and reasoned with. To that end, we gather a diverse range of characters and manually create questions that systematically evaluate various aspects of role knowledge. Such created questions usually demand a thorough understanding and adaptable use of role knowledge, as well as multi-level reasoning over role knowledge.

\subsection{Character Collection}
RoleEval compiles its character set from diverse sources, including Wikipedia\footnote{\url{https://www.wikipedia.org}}, Baidu Baike\footnote{\url{https://baike.baidu.com}}, Fandom\footnote{\url{https://www.fandom.com}}, and Moegirlpedia.\footnote{\url{https://zh.moegirl.org.cn}} While Wikipedia and Baidu Baike provide a wide range of information, Fandom and Moegirlpedia offer detailed insights into anime, comics, and games. For \textit{RoleEval-Global}, we gather 200 characters known internationally across five categories: celebrities, anime/comics, movies/TV series, games, and fiction, ensuring diversity by selecting an equal number of characters from each category and different works for each fictional character. \textit{RoleEval-Chinese} extends this collection with an additional 100 characters popular primarily in China, totaling 300 characters for RoleEval. These characters are detailed in the appendix \ref{sec:character_list}.

Before manually creating questions related to collected characters, we rigorously verify each character's encyclopedia information to ensure comprehensiveness. We prefer characters with abundant, collaboratively-contributed information for its reliability and objectivity, ensuring our dataset's diversity, accuracy, and true representation of characters. Additionally, we assess character popularity on social media (via follower counts and engagement) and search engine presence (by search result volume). During the manual question creation phase, we refer to multiple online encyclopedias, selecting only consistent knowledge points to avoid conflicts.

\begin{table}[t]
    \centering
    \Large
    \resizebox{\linewidth}{!}{%
        \renewcommand*{\arraystretch}{1.1}
        \begin{tabular}{@{}c|ccc@{}}
        \toprule
        \multicolumn{1}{l|}{}                      & \textbf{RoleEval-Global} & \textbf{RoleEval-Chinese} & \textbf{Total}   \\ \midrule
        \textbf{\# Characters}                      & 200                      & 100                       & 300              \\
        \textbf{\# Questions}                       & 4,000                    & 2,000                     & 6,000            \\
        \textbf{Languages}                         & zh, en         & zh, en          & zh, en \\
        \textbf{\makecell{Avg. Q. Tokens (zh)}} & 24.6                     & 26.1                      & 25.1             \\
        \textbf{\makecell{Avg. Q. Tokens (en)}} & 14.5                     & 16.0                      & 15.0             \\ \bottomrule
        \end{tabular}%
    }
    \caption{Statistics of RoleEval. Here we use zh and en to indicate Chinese and English respectively.}
    \label{tab:basic_information}
    
\end{table}

\subsection{Question Design}
Our focus is primarily on factual knowledge and comprehensively assessing the ability of LLMs in capturing and understanding various character roles and contexts. In addition to basic knowledge that is directly stated in encyclopedias, we also design multi-hop questions to examine the ability of LLMs to dynamically combine and reason with existing knowledge.

Specifically, after collecting the 300 characters from online encyclopedias, we build RoleEval in the form of multiple-choice questions, with four options for each question, which is a common practice adopted by many existing benchmarks such as MMLU \citep{hendrycksMeasuringMassiveMultitask2021} and C-Eval \citep{huangCEvalMultiLevelMultiDiscipline2023}. Each character is associated with 17 and 3 unique questions for basic knowledge and multi-hop reasoning respectively, thus culminating in a total of 6,000 questions. The statistics of RoleEval are shown in Table \ref{tab:basic_information}. We also illustrate the length distribution of created questions in Figure \ref{fig:RoleEval_length}.

In RoleEval, we consider three types of fundamental knowledge required to depict a character:
\begin{enumerate}
    \item \textbf{Inherent Attributes}: This type of knowledge includes the fundamental characteristics intrinsic to the character, such as gender, race, personality, skills, and abilities. These attributes are typically presented in a tabular format, or directly described in online encyclopedias.
    \item \textbf{Social Relationships}: This type of knowledge pertains to the relationships of the character with other individuals, which could include parents, disciples, and other significant personal or professional relationships.
    \item \textbf{Experiences}: This type of knowledge details the experiences or events that the character has undergone. For real-world individuals, this usually includes significant life events or experiences in which they were direct participants. For fictional characters, this involves extracting key plot points or story arcs described in online encyclopedias.
\end{enumerate}

\begin{figure*}[ht]
    \centering
    \includegraphics[width=\linewidth]{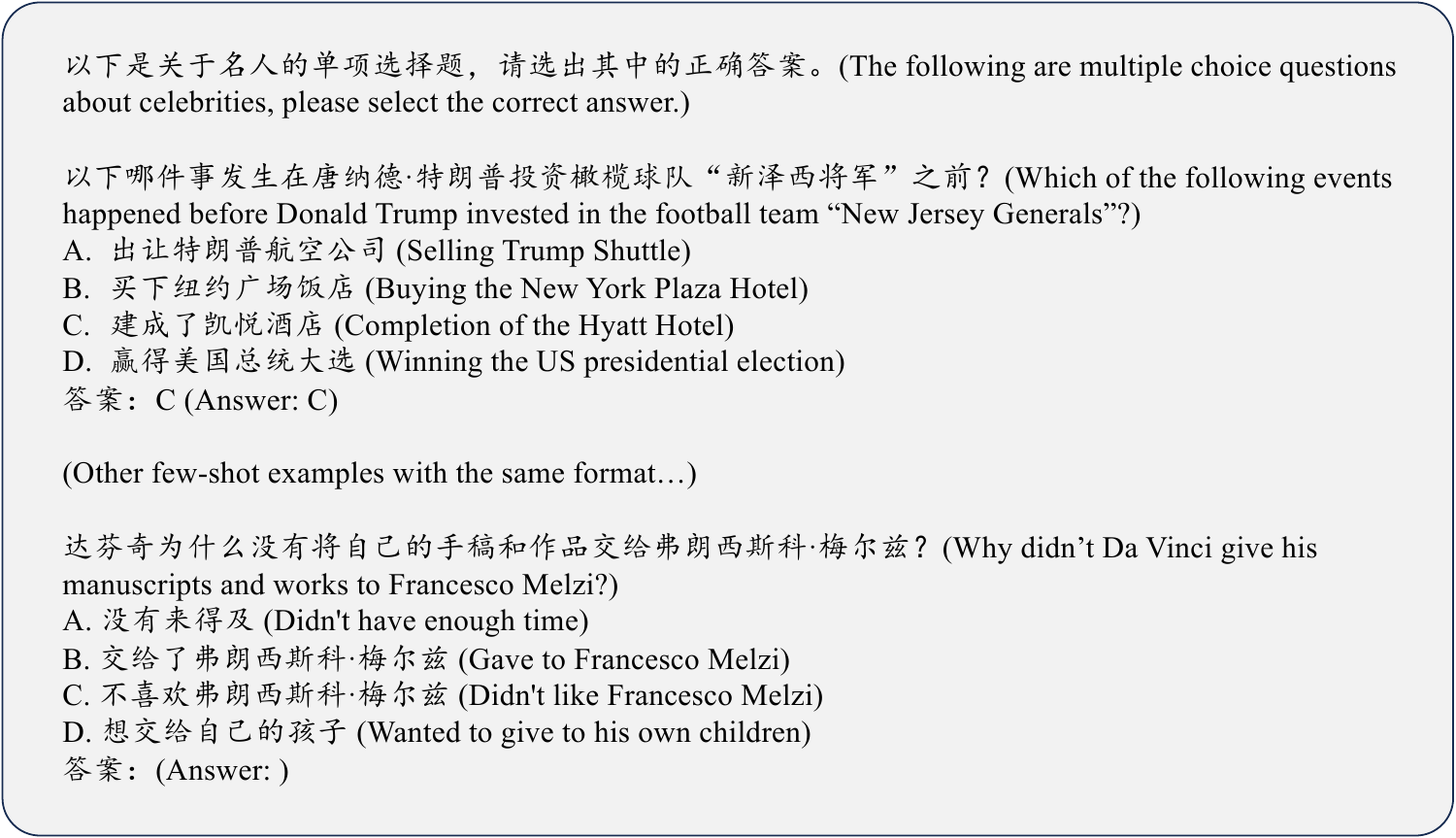}
    \caption{Few-shot prompt examples for RoleEval. English translations are placed after the corresponding Chinese text for clarity.}
    \label{fig:question_example}
    
\end{figure*}

To enhance the comprehensive assessment and diversify the scope of multiple-choice questions, our approach extends beyond merely querying knowledge from online encyclopedias. As a supplement to direct questions, we incorporate two additional question types, which are designed to be combined with the previously identified four types of knowledge for a more dynamic and comprehensive evaluation:

\begin{enumerate}
    \item \textbf{Negation Question Type:} These questions, usually formatted as ``Which of the following is NOT...'', require a comprehensive understanding of a specific knowledge point.%

    \item \textbf{Non-occurrence Scenario Question Type:} These questions test for non-occurrences, with correct answers often framed in the negative (e.g., ``Did not happen...''). This format examines whether the model generates illusions or false assumptions.%
\end{enumerate}

We further add three reasoning questions for each character that need multi-hop reasoning over these types of fundamental knowledge. According to the knowledge required in the intermediate reasoning steps, we classify these reasoning questions into three categories, and assign one question for each available (role, reasoning category) pair:

\begin{enumerate}
    \item \textbf{Character Relationship Reasoning:} When answering this type of questions, models need to reason about the relationship between characters.%
    \item \textbf{Event Participant Reasoning:} To answer a question in this category, models need to reason out the participants of an event, and then combine it with other information in the question to locate the answer. %
    \item \textbf{Timeline Reasoning:} Answering questions in this group requires models to understand the sequence of events, infer the time of occurrence of the event in question stems and options, and then select the correct option based on the timeline.%
\end{enumerate}

These reasoning questions go beyond simply memorizing one-hop knowledge in text, intending to connect multiple related characters, events, and storylines. It requires the compositionality capability of models to intrinsically and dynamically combine multiple knowledge points and answer the given questions, thus making our RoleEval a challenging benchmark for LLMs. For more detailed information, we illustrate the examples of negation, non-occurrence and different types of reasoning in Appendix \ref{sec:example_negation_non_occurrence_questions} and \ref{sec:example_reasoning_questions}.

\subsection{Quality Check}
To ensure quality and boost efficiency during benchmark construction, we propose a hybrid quality check process with the combination of both automatic and human verification. In the automatic checking stage, we use GPT-4 and GPT-3.5 API\footnote{Unless otherwise specified, the term ``GPT-4'' and ``GPT-3.5'' shall henceforth be used to represent \texttt{gpt-4-0613} and \texttt{gpt-3.5-turbo-0613} respectively.} to control the degree of difficulty and discrimination ability. Assume the accuracy of GPT-4 and GPT-3.5 API for answering questions of character $c$ is $x_c$ and $y_c$ respectively. We apply the below criteria as the preliminary and instant feedback for human annotators:

\begin{gather}
  x_l \leq x_c \leq x_u \label{eq1} \\
  y_c \leq y_u \\
  x_c - y_c \geq d \label{eq3}
\end{gather}

where $x_l$ and $x_u$ indicate the predefined lower threshold and upper threshold for $x_c$, and $y_u$ indicates the predefined upper threshold for $y_c$. These three hyperparameters control the overall degree of difficulty, making sure that questions are not too easy or too hard for LLMs, in terms of the answering accuracy of GPT-4/3.5. And $d$ is the lower threshold of difference between $x_c$ and $y_c$. Since GPT-4 is a significantly more powerful model than GPT-3.5 in practice, we use this hyperparameter to ensure the discrimination ability of this benchmark for various models. Only questions that satisfy the conditions in Eq. (\ref{eq1})-(\ref{eq3}) are kept. In our preliminary study, we find $x_l = 0.3$, $x_u = 0.9$, $y_u = 0.8$, and $d = 0.15$ achieve appropriate difficulty and discrimination ability control for questions.

After this automatic scrutinization, we manually check the kept questions and options to ensure the quality of our benchmark. To easily check the factual correctness and prevent questions from overemphasis on peripheral aspects, we ask annotators to also provide links to the referenced text in the online encyclopedias along with each question to achieve effective oversight.

\subsection{Translation}

To evaluate LLMs in different languages, we translate Chinese RoleEval questions to English using GPT-4.\footnote{We use RoleEval (\textbf{en}) and RoleEval (\textbf{zh}) to indicate RoleEval in English and Chinese respectively.} We have found that GPT-4 produce more adaptable and precise character-specific translations than traditional machine translation engines like Google Translate, due to its customizable prompts. To address potential ambiguities in entity translations, we hire human translators to review and adjust them based on the original Chinese content. Given RoleEval's focus on factual accuracy over stylistic elements, combining machine translation with human post-editing effectively minimizes information loss.

\section{Experiments}
We evaluated a wide range of English and Chinese LLMs on RoleEval, aiming to analyze the capturing, utilizing, and reasoning capabilities of role knowledge for these LLMs. 

\begin{table}[t]
    \HUGE
    \centering
    \resizebox{0.5\textwidth}{!}{%
    \renewcommand*{\arraystretch}{1.2}
    \begin{tabular}{cccc}
    \hline
        \textbf{Models} & \textbf{Open Source?} & \textbf{Model Size} & \textbf{Primary Language} \\ \hline
        BLOOM & Yes & 1.1B, 1.7B, 3B, 7.1B & English \\ 
        Pythia & Yes & 1.4B, 2.8B, 6.9B, 12B & English \\ 
        LLaMA & Yes & 7B, 13B, 30B, 65B & English \\ 
        LLaMA-2 & Yes & 7B, 13B, 70B & English \\ 
        Falcon & Yes & 7B, 40B & English \\ 
        Mistral & Yes & 7B & English \\ 
        GPT-4 & No & undisclosed & English \\ 
        GPT-3.5 & No & undisclosed & English \\
        \hline
        ChatGLM3 & Yes & 6B & Chinese \\ 
        Baichuan2 & Yes & 7B, 13B & Chinese \\ 
        Qwen & Yes & 1.8B, 7B, 14B, 72B & Chinese \\ 
        Chinese-LLaMA-2 & Yes & 7B, 13B & Chinese \\ 
        Skywork & Yes & 13B & Chinese \\ 
        Yi & Yes & 6B, 34B & Chinese \\ 
        MiniMax & No & undisclosed & Chinese \\ \hline
    \end{tabular}
    }
    \caption{Details of models evaluated in our experiments. For simplicity, we only list the sizes of evaluated models in the ``Model Size'' column.}
    \label{tab:model_list}
    
\end{table}

\begin{table*}[ht]
\large
\centering
\resizebox{\textwidth}{!}{%
\renewcommand*{\arraystretch}{1.1}
\begin{tabular}{@{}c|cccccc|cccccc@{}}
\toprule
\multirow{2}{*}{\textbf{Model}} & \multicolumn{6}{c|}{\textbf{RoleEval-Global (4,000 questions)}}                                                                            & \multicolumn{6}{c}{\textbf{RoleEval-Chinese (2,000 questions)}}                                                                            \\ \cmidrule(l){2-13} 
                                & \textbf{CE}    & \textbf{AC}    & \textbf{MT}    & \textbf{GA}    & \multicolumn{1}{c|}{\textbf{FI}}    & \textbf{Avg.}  & \textbf{CE}    & \textbf{AC}    & \textbf{MT}    & \textbf{GA}    & \multicolumn{1}{c|}{\textbf{FI}}    & \textbf{Avg.}  \\ \midrule
GPT-3.5-0613                    & 46.62          & 48.38          & 51.75          & 49.50          & \multicolumn{1}{c|}{47.38}          & 48.73          & 42.25          & 43.50          & 39.75          & 43.75          & \multicolumn{1}{c|}{39.00}          & 41.65          \\
GPT-3.5-1106                    & 48.75          & 51.88          & 51.25          & 49.88          & \multicolumn{1}{c|}{48.38}          & 50.02          & 47.50          & 46.75          & 41.75          & 44.75          & \multicolumn{1}{c|}{38.75}          & 43.90          \\
GPT-4-0613                      & 73.38          & 72.12          & 74.25          & 72.25          & \multicolumn{1}{c|}{69.62}          & 72.32          & 57.75          & 60.25          & 57.75          & 60.00          & \multicolumn{1}{c|}{58.00}          & 58.75          \\
GPT-4-1106                      & \textbf{74.75} & \textbf{73.62} & \textbf{74.38} & \textbf{72.50} & \multicolumn{1}{c|}{71.62}          & \textbf{73.38} & 62.50          & \textbf{63.25} & 63.00          & \textbf{62.00} & \multicolumn{1}{c|}{63.00}          & 62.75          \\
Falcon-7B                       & 23.88          & 28.12          & 24.50          & 28.00          & \multicolumn{1}{c|}{28.12}          & 26.52          & 24.75          & 30.50          & 31.50          & 29.75          & \multicolumn{1}{c|}{25.25}          & 28.35          \\
Falcon-40B                      & 39.62          & 32.25          & 32.38          & 30.00          & \multicolumn{1}{c|}{45.00}          & 35.85          & 28.25          & 33.00          & 30.25          & 29.25          & \multicolumn{1}{c|}{38.50}          & 31.85          \\
LLaMA-7B                        & 25.50          & 31.87          & 25.87          & 26.00          & \multicolumn{1}{c|}{28.88}          & 27.62          & 28.50          & 24.75          & 20.50          & 27.75          & \multicolumn{1}{c|}{29.00}          & 26.10          \\
LLaMA-13B                       & 28.50          & 28.50          & 28.25          & 26.50          & \multicolumn{1}{c|}{27.75}          & 27.90          & 27.25          & 29.75          & 27.25          & 26.00          & \multicolumn{1}{c|}{29.00}          & 27.85          \\
LLaMA-30B                       & 24.88          & 31.13          & 30.25          & 27.75          & \multicolumn{1}{c|}{28.62}          & 28.52          & 30.00          & 28.75          & 26.00          & 31.75          & \multicolumn{1}{c|}{28.00}          & 28.90          \\
LLaMA-65B                       & 32.12          & 31.87          & 32.75          & 31.00          & \multicolumn{1}{c|}{34.88}          & 32.52          & 30.00          & 32.25          & 29.00          & 35.50          & \multicolumn{1}{c|}{29.00}          & 31.15          \\
LLaMA-2-7B                      & 37.00          & 29.88          & 28.75          & 34.50          & \multicolumn{1}{c|}{38.25}          & 33.67          & 25.75          & 28.00          & 33.75          & 29.75          & \multicolumn{1}{c|}{34.50}          & 30.35          \\
LLaMA-2-13B                     & 36.50          & 34.00          & 33.00          & 31.87          & \multicolumn{1}{c|}{31.75}          & 33.42          & 28.75          & 30.50          & 25.25          & 29.75          & \multicolumn{1}{c|}{28.25}          & 28.50          \\
LLaMA-2-70B                     & 53.50          & 43.25          & 39.25          & 40.25          & \multicolumn{1}{c|}{47.25}          & 44.70          & 36.00          & 38.00          & 36.25          & 36.25          & \multicolumn{1}{c|}{34.75}          & 36.25          \\
Mistral-7B                      & 36.12          & 33.50          & 32.00          & 30.25          & \multicolumn{1}{c|}{35.00}          & 33.38          & 32.50          & 37.50          & 26.25          & 33.25          & \multicolumn{1}{c|}{31.50}          & 32.20          \\ \midrule
MiniMax                         & 51.75          & 54.50          & 62.62          & 56.75          & \multicolumn{1}{c|}{52.75}          & 55.67          & 54.00          & 55.00          & 52.75          & 57.50          & \multicolumn{1}{c|}{54.00}          & 54.65          \\
Baichuan2-7B                    & 56.00          & 49.62          & 45.50          & 40.50          & \multicolumn{1}{c|}{52.38}          & 48.80          & 52.25          & 43.75          & 49.00          & 47.25          & \multicolumn{1}{c|}{55.00}          & 49.45          \\
Baichuan2-13B                   & 60.25          & 52.38          & 51.00          & 46.88          & \multicolumn{1}{c|}{60.75}          & 54.25          & 54.75          & 47.75          & 54.00          & 47.50          & \multicolumn{1}{c|}{60.00}          & 52.80          \\
ChatGLM3-6B                     & 56.50          & 47.62          & 48.38          & 41.88          & \multicolumn{1}{c|}{54.50}          & 49.78          & 50.00          & 44.50          & 48.00          & 44.25          & \multicolumn{1}{c|}{58.00}          & 48.95          \\
Chinese-LLaMA-2-7B              & 35.62          & 36.75          & 35.62          & 35.38          & \multicolumn{1}{c|}{34.38}          & 35.55          & 34.50          & 29.00          & 33.00          & 30.25          & \multicolumn{1}{c|}{36.25}          & 32.60          \\
Chinese-LLaMA-2-13B             & 45.38          & 38.25          & 39.88          & 31.87          & \multicolumn{1}{c|}{42.12}          & 39.50          & 36.50          & 36.50          & 34.00          & 34.00          & \multicolumn{1}{c|}{40.50}          & 36.30          \\
Qwen-7B                         & 54.75          & 44.38          & 44.62          & 42.75          & \multicolumn{1}{c|}{53.00}          & 47.90          & 49.00          & 42.00          & 47.50          & 44.75          & \multicolumn{1}{c|}{51.25}          & 46.90          \\
Qwen-14B                        & 62.50          & 52.38          & 55.00          & 45.50          & \multicolumn{1}{c|}{58.00}          & 54.67          & 56.25          & 45.50          & 54.75          & 51.50          & \multicolumn{1}{c|}{56.75}          & 52.95          \\
Qwen-72B                        & 72.88          & 63.88          & 70.38          & 56.75          & \multicolumn{1}{c|}{\textbf{73.50}} & 67.47          & \textbf{70.00} & 59.75          & 66.00          & 61.25          & \multicolumn{1}{c|}{74.00}          & \textbf{66.20} \\
Skywork-13B                     & 59.13          & 51.75          & 51.88          & 44.50          & \multicolumn{1}{c|}{58.75}          & 53.20          & 55.25          & 45.75          & 56.00          & 48.50          & \multicolumn{1}{c|}{57.50}          & 52.60          \\
Yi-6B                           & 61.88          & 51.38          & 52.38          & 45.38          & \multicolumn{1}{c|}{60.75}          & 54.35          & 59.25          & 46.00          & 61.50          & 47.75          & \multicolumn{1}{c|}{62.00}          & 55.30          \\
Yi-34B                          & 72.38          & 60.62          & 69.75          & 53.25          & \multicolumn{1}{c|}{73.12}          & 65.83          & 65.50          & 54.50          & \textbf{70.00} & 56.00          & \multicolumn{1}{c|}{\textbf{77.00}} & 64.60          \\ \bottomrule
\end{tabular}%
}
\caption{Five-shot results on RoleEval (\textbf{zh}) in five categories: celebrities (\textbf{CE}), anime and comics (\textbf{AC}), movie and TV series (\textbf{MT}), games (\textbf{GA}), and fiction (\textbf{FI}).}
\label{tab:zh_fewshot_results}

\end{table*}

\subsection{Setup}
We implemented an evaluation pipeline for RoleEval with the lm-evaluation-harness framework \citep{gaoFrameworkFewshotLanguage2023} under both zero- and five-shot setting. Considering RoleEval and MMLU \citep{hendrycksMeasuringMassiveMultitask2021} share the same multiple-choice format, we adopted a similar setup with MMLU for RoleEval. Specifically, for open-source models, we calculated the probability of subsequent tokens following the initial prompt. Among the options ``A'', ``B'', ``C'', and ``D'', we chose the one with the highest probability as the preferred choice of the model. For closed-source models such as GPT-4 \citep{openaiGPT4TechnicalReport2023}, we followed \citet{huangCEvalMultiLevelMultiDiscipline2023} and \citet{liCMMLUMeasuringMassive2023} to use regular expressions to extract the preferred choice of the model. All experiments in this paper were conducted using two NVIDIA A800 80G GPUs.

\subsection{Prompt}
Figure \ref{fig:question_example} illustrates the prompt we used for evaluation. For each question, we added ``\chin{以下是关于[category]的单项选择题，请选出其中的正确答案。}'' (``The following multiple-choice questions are about [category]. Please choose the correct answer.'') before the question stem, and ``\chin{答案：}'' (``Answer: '') after four options, where ``[category]'' was chosen from \chin{名人} (``celebrities''), \chin{动漫角色} (``anime and comics''), \chin{影视角色} (``movies and TV series''), \chin{游戏角色} (``games'') and \chin{小说人物} (``fictions''). For the five-shot setting, we added in-context examples before the actual question to answer. These examples shared the same format as the actual question, except that the ground truth option was provided for each in-context example.

\begin{table*}[h]
\large
\centering
\resizebox{\textwidth}{!}{%
\renewcommand*{\arraystretch}{1.1}
\begin{tabular}{@{}c|cccccc|cccccc@{}}
\toprule
\multirow{2}{*}{\textbf{Model}} & \multicolumn{6}{c|}{\textbf{RoleEval-Global   (4,000 questions)}}                                                        & \multicolumn{6}{c}{\textbf{RoleEval-Chinese   (2,000 questions)}}                                                        \\ \cmidrule(l){2-13} 
                                & \textbf{CE}    & \textbf{AC}    & \textbf{MT}    & \textbf{GA}    & \multicolumn{1}{c|}{\textbf{FI}}    & \textbf{Avg.}  & \textbf{CE}    & \textbf{AC}    & \textbf{MT}    & \textbf{GA}    & \multicolumn{1}{c|}{\textbf{FI}}    & \textbf{Avg.}  \\ \midrule
GPT-3.5-0613                    & 57.38          & 59.62          & 58.13          & 59.50          & \multicolumn{1}{c|}{57.50}          & 58.43          & 42.00          & 47.75          & 42.50          & 42.25          & \multicolumn{1}{c|}{45.50}          & 44.00          \\
GPT-3.5-1106                    & 58.75          & 56.62          & 55.75          & 58.00          & \multicolumn{1}{c|}{55.00}          & 56.82          & 38.25          & 45.50          & 44.00          & 44.50          & \multicolumn{1}{c|}{46.00}          & 43.65          \\
GPT-4-0613                      & \textbf{77.62} & \textbf{79.50} & 73.12          & 74.88          & \multicolumn{1}{c|}{\textbf{75.00}} & \textbf{76.02} & 54.25          & 61.75          & \textbf{63.00} & \textbf{63.00} & \multicolumn{1}{c|}{\textbf{63.00}} & \textbf{61.00} \\
GPT-4-1106                      & 75.12          & 78.75          & \textbf{75.00} & \textbf{76.12} & \multicolumn{1}{c|}{\textbf{75.00}} & 76.00          & \textbf{57.50} & \textbf{63.50} & 60.00          & 62.50          & \multicolumn{1}{c|}{58.00}          & 60.30          \\
Falcon-7B                       & 26.25          & 27.75          & 28.50          & 29.38          & \multicolumn{1}{c|}{31.00}          & 28.58          & 27.25          & 27.75          & 27.75          & 29.75          & \multicolumn{1}{c|}{28.50}          & 28.20          \\
Falcon-40B                      & 47.38          & 45.00          & 49.62          & 43.12          & \multicolumn{1}{c|}{50.00}          & 47.02          & 34.00          & 38.25          & 30.75          & 38.75          & \multicolumn{1}{c|}{35.25}          & 35.40          \\
LLaMA-7B                        & 29.38          & 30.50          & 29.25          & 33.50          & \multicolumn{1}{c|}{28.50}          & 30.23          & 24.00          & 27.50          & 29.75          & 33.00          & \multicolumn{1}{c|}{29.25}          & 28.70          \\
LLaMA-13B                       & 39.38          & 40.25          & 39.88          & 40.62          & \multicolumn{1}{c|}{43.00}          & 40.63          & 32.75          & 31.75          & 30.75          & 38.50          & \multicolumn{1}{c|}{32.00}          & 33.15          \\
LLaMA-30B                       & 51.62          & 46.88          & 48.62          & 43.12          & \multicolumn{1}{c|}{52.62}          & 48.57          & 34.75          & 35.75          & 30.75          & 40.00          & \multicolumn{1}{c|}{35.00}          & 35.25          \\
LLaMA-65B                       & 58.13          & 50.50          & 54.37          & 47.62          & \multicolumn{1}{c|}{54.50}          & 53.02          & 41.50          & 38.50          & 33.50          & 43.25          & \multicolumn{1}{c|}{37.50}          & 38.85          \\
LLaMA-2-7B                      & 38.88          & 37.00          & 37.50          & 41.62          & \multicolumn{1}{c|}{42.38}          & 39.48          & 28.75          & 29.25          & 32.75          & 37.50          & \multicolumn{1}{c|}{32.25}          & 32.10          \\
LLaMA-2-13B                     & 49.38          & 43.50          & 46.50          & 44.25          & \multicolumn{1}{c|}{48.25}          & 46.38          & 30.50          & 36.50          & 33.25          & 36.50          & \multicolumn{1}{c|}{33.25}          & 34.00          \\
LLaMA-2-70B                     & 63.25          & 57.38          & 59.00          & 50.00          & \multicolumn{1}{c|}{63.25}          & 58.58          & 43.25          & 41.50          & 40.25          & 47.50          & \multicolumn{1}{c|}{43.50}          & 43.20          \\
Mistral-7B                      & 54.87          & 46.75          & 49.62          & 44.25          & \multicolumn{1}{c|}{52.25}          & 49.55          & 35.75          & 42.00          & 30.00          & 41.75          & \multicolumn{1}{c|}{31.50}          & 36.20          \\ \midrule
MiniMax                         & 54.87          & 56.38          & 53.50          & 54.12          & \multicolumn{1}{c|}{51.38}          & 54.05          & 34.00          & 39.50          & 40.75          & 38.25          & \multicolumn{1}{c|}{39.00}          & 38.30          \\
Baichuan2-7B                    & 51.00          & 45.12          & 49.00          & 42.12          & \multicolumn{1}{c|}{50.00}          & 47.45          & 37.25          & 35.75          & 33.00          & 40.25          & \multicolumn{1}{c|}{37.00}          & 36.65          \\
Baichuan2-13B                   & 56.12          & 47.50          & 51.50          & 45.62          & \multicolumn{1}{c|}{54.00}          & 50.95          & 35.50          & 36.50          & 31.25          & 42.25          & \multicolumn{1}{c|}{34.75}          & 36.05          \\
ChatGLM3-6B                     & 55.12          & 46.62          & 49.25          & 43.25          & \multicolumn{1}{c|}{52.62}          & 49.37          & 36.25          & 36.25          & 35.25          & 42.25          & \multicolumn{1}{c|}{43.50}          & 38.70          \\
Chinese-LLaMA-2-13B             & 47.75          & 46.00          & 46.88          & 45.00          & \multicolumn{1}{c|}{48.38}          & 46.80          & 34.00          & 38.50          & 27.75          & 37.50          & \multicolumn{1}{c|}{34.00}          & 34.35          \\
Chinese-LLaMA-2-7B              & 36.50          & 30.75          & 31.75          & 36.25          & \multicolumn{1}{c|}{39.50}          & 34.95          & 30.50          & 27.75          & 33.00          & 30.50          & \multicolumn{1}{c|}{27.75}          & 29.90          \\
Qwen-7B                         & 53.87          & 46.12          & 48.12          & 40.00          & \multicolumn{1}{c|}{51.12}          & 47.85          & 36.25          & 36.00          & 36.25          & 42.25          & \multicolumn{1}{c|}{40.00}          & 38.15          \\
Qwen-14B                        & 61.12          & 49.00          & 53.87          & 45.38          & \multicolumn{1}{c|}{56.12}          & 53.10          & 41.00          & 38.75          & 38.25          & 43.25          & \multicolumn{1}{c|}{41.00}          & 40.45          \\
Qwen-72B                        & 70.12          & 62.00          & 69.00          & 55.75          & \multicolumn{1}{c|}{69.50}          & 65.27          & 52.75          & 47.50          & 46.50          & 54.25          & \multicolumn{1}{c|}{50.50}          & 50.30          \\
Skywork-13B                     & 56.25          & 46.75          & 51.62          & 44.38          & \multicolumn{1}{c|}{53.62}          & 50.52          & 39.25          & 34.50          & 38.25          & 41.75          & \multicolumn{1}{c|}{38.50}          & 38.45          \\
Yi-6B                           & 59.25          & 52.00          & 54.12          & 47.50          & \multicolumn{1}{c|}{56.25}          & 53.82          & 42.25          & 38.50          & 41.50          & 44.25          & \multicolumn{1}{c|}{45.00}          & 42.30          \\
Yi-34B                          & 73.12          & 61.75          & 67.88          & 57.12          & \multicolumn{1}{c|}{67.25}          & 65.42          & 56.00          & 52.00          & 47.50          & 55.00          & \multicolumn{1}{c|}{57.00}          & 53.50          \\ \bottomrule
\end{tabular}%
}
\caption{Five-shot results on RoleEval (\textbf{en}) in five categories: celebrities (\textbf{CE}), anime and comics (\textbf{AC}), movie and TV series (\textbf{MT}), games (\textbf{GA}), and fiction (\textbf{FI}).}
\label{tab:en_fewshot_results}

\end{table*}

\subsection{Models}

We selected a wide range of publicly available LLMs for evaluation. Due to the limit of computational resources, we only evaluated models with the number of parameters ranging from 1B to 72B parameters since they can produce meaningful results and be loaded in two 80GB Nvidia A800 GPUs with bf16 format.

\paragraph{Chinese LLMs} For Chinese questions in RoleEval, we evaluated popular Chinese open-source LLMs, such as ChatGLM \citep{duGLMGeneralLanguage2022, zengGLM130BOpenBilingual2023}, Baichuan \citep{yangBaichuanOpenLargescale2023}, Qwen \citep{baiQwenTechnicalReport2023}, Yi\footnote{\url{https://github.com/01-ai/Yi}}, Skywork \citep{weiSkyworkMoreOpen2023}, Chinese-LLaMA-2 \citep{cuiEfficientEffectiveText2023a}, along with close-sourced LLMs like Minimax.\footnote{\url{https://api.minimax.chat}}

\paragraph{English LLMs} For English questions in RoleEval, we evaluated open-source LLMs including BLOOM \citep{workshopBLOOM176BParameterOpenAccess2023}, Pythia \citep{bidermanPythiaSuiteAnalyzing2023}, LLaMA \citep{touvronLLaMAOpenEfficient2023}, LLaMA2 \citep{touvronLlamaOpenFoundation2023}, Falcon \citep{almazrouei2023falcon}, Mistral-7B \citep{jiangMistral7B2023}, and two closed-source LLMs: ChatGPT \citep{ouyangTrainingLanguageModels2022} and GPT-4 \citep{openaiGPT4TechnicalReport2023}.

Since Baichuan, Qwen, Yi, Chinese-LLaMA-2, BLOOM, Pythia, LLaMA, LLaMA2, and Falcon have multiple sizes of LLMs that satisfy our model restriction, we evaluated various sizes of LLMs for each LLM family as shown in Table \ref{tab:model_list}.

\subsection{Results and Analysis}

\paragraph{Overall Performance} Table \ref{tab:zh_fewshot_results} and \ref{tab:en_fewshot_results} show the few-shot experimental results. Since the zero-shot results are generally lower than the few-shot experimental results, we provide them in Appendix \ref{sec:zeroshot_results}. We find that GPT-4 maintains a lead on \textit{RoleEval-Global}, with its latest version (\texttt{gpt-4-1106}) outperforming the earlier (\texttt{gpt-4-0613}). We believe this superiority is attributed to the more recent knowledge cutoff in \texttt{gpt-4-1106}, enhancing its performance in domains with rapidly evolving information. Nevertheless, the overall accuracy still indicates large room for improvement even for the state-of-the-art LLMs. On the \textit{RoleEval-Chinese} (\textbf{zh}) dataset, some Chinese LLMs, such as Qwen-72B and Yi-34B, show superior performance to GPT-4. This is likely due to their higher proportion of Chinese training data and abundant high-quality discussions about characters in \textit{RoleEval-Chinese} on Chinese online platforms. Notably, GPT-4 retains its edge in anime, comics, and games, where many characters are also popular in the English-speaking world. These results highlight the importance of choosing balanced training data and evaluating role knowledge across various languages and cultural settings. We also provide results and analysis grouped by knowledge and reasoning questions in Appendix \ref{sec:results_by_knowledge_and_reasoning}.

\paragraph{Differences in Languages} We observe a significant improvement in GPT-4 and 3.5's performance on RoleEval (\textbf{en}) dataset compared to RoleEval (\textbf{zh}), even though these two parts of the dataset have identical semantic content. This observation is also evident in predominantly English-language open-source models, especially for models with little Chinese training data, such as LLaMA, Mistral, and Falcon. This suggests that even the most powerful LLMs still lack effective cross-lingual knowledge transfer, indicating that these models fail to build complete bi-directional mappings between entities in different languages. Conversely, Chinese models generally underperform on the English RoleEval datasets, highlighting a similar language-specific limitation, though the gap is narrower since Chinese LLMs still use a large amount of English pre-training data. Among all open-source models, Yi-34B achieves almost the same performance in both Chinese and English on \textit{RoleEval-Global}, indicating its balanced training for global influential characters in both languages.

\paragraph{Comparative Analysis of Open-Source Models} LLaMA2-70B has emerged as the best open-source model primarily trained in English, closely matching GPT-3.5. While for Chinese LLMs, Qwen-72B and Yi-34B not only surpass GPT-3.5 but also exceed GPT-4 on the \textit{RoleEval-Chinese} (\textbf{zh}) dataset. However, these Chinese models still show noticeable gaps compared to GPT-4 in other scenarios.

\begin{figure}[t]
    \centering
    \includegraphics[width=\linewidth]{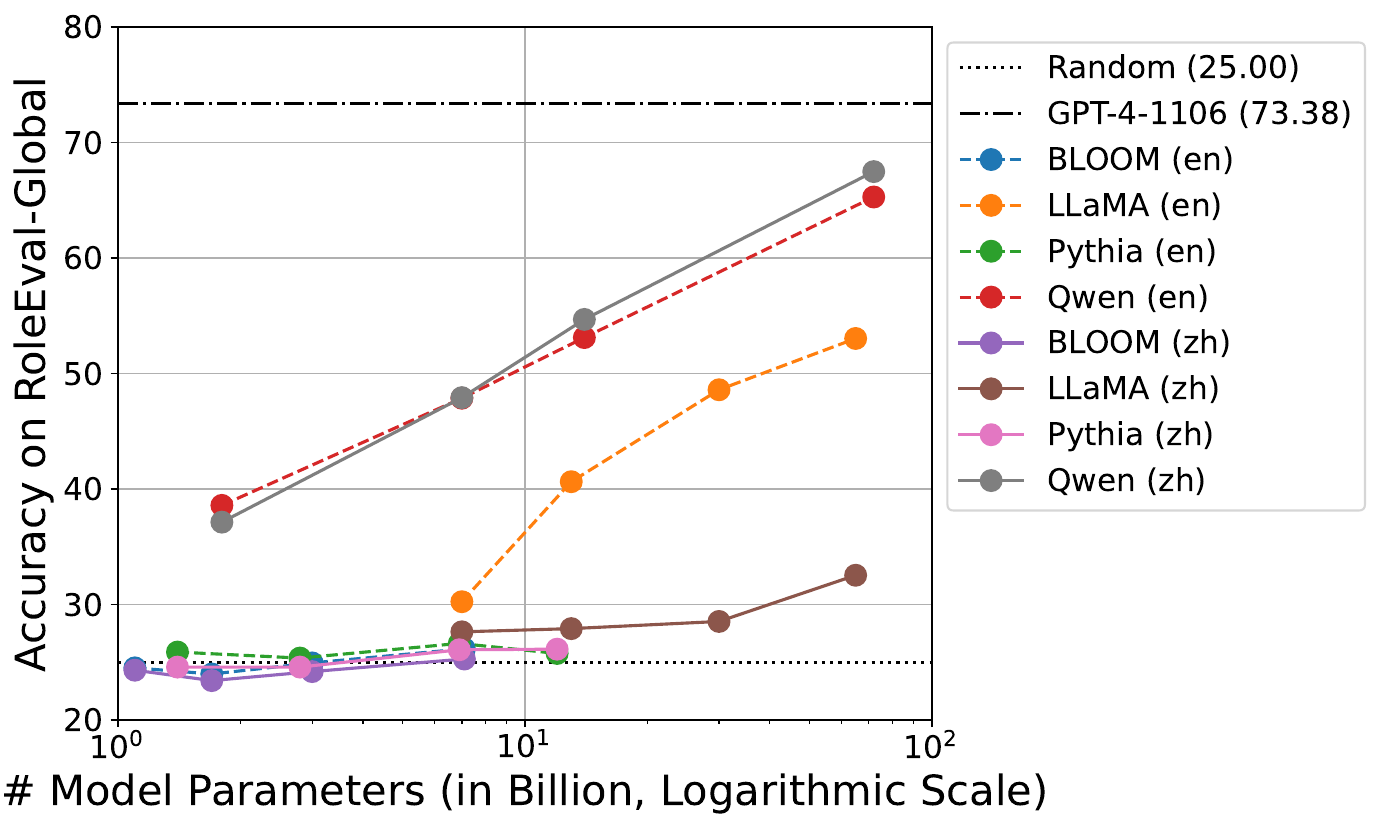}
    \caption{Parameter Scaling Laws on \textit{RoleEval-Global}.}
    \label{fig:param_scaling}
    
\end{figure}

\paragraph{Parameter Scaling Laws} We also analyzed the correlation between the accuracy of role knowledge and the size and the number of training tokens of LLMs on \textit{RoleEval-Global}.\footnote{To obtain meaningful results, we only chose model families with more than three models in this experiment, and we required the amount of training data and training tokens to be the same for models with different parameters within the same model family.} As shown in Figure \ref{fig:param_scaling}, accuracy generally improves with model size for LLaMA and Qwen, and the trend is consistent with previously established knowledge transfer patterns: For Chinese LLMs, the rate of improvement is greater for RoleEval (\textbf{zh}) than for RoleEval (\textbf{en}). In contrast, LLMs primarily trained on English corpora show opposite trends. Notably, BLOOM and Pythia do not show performance improvements across various settings. We speculate that this is due to their relatively lower token training volume (366B for BLOOM and 300B for Pythia), while most models with great performance have already been trained on more than 1TB tokens.

\paragraph{Token Scaling Laws} To further explore the scaling law on the number of tokens, we conducted experiments on publicly available intermediate checkpoints of BLOOM-7B1, Pythia-6.9B, Baichuan-7B, and Skywork-13B. For BLOOM, Baichuan, and Skywork, we selected all available intermediate checkpoints, resulting in 8, 11, and 6 checkpoints respectively. For Pythia, we chose a checkpoint every 13,000 steps and obtained 12 intermediate checkpoints for evaluation.

\begin{figure}[t]
    \centering
    \includegraphics[width=\linewidth]{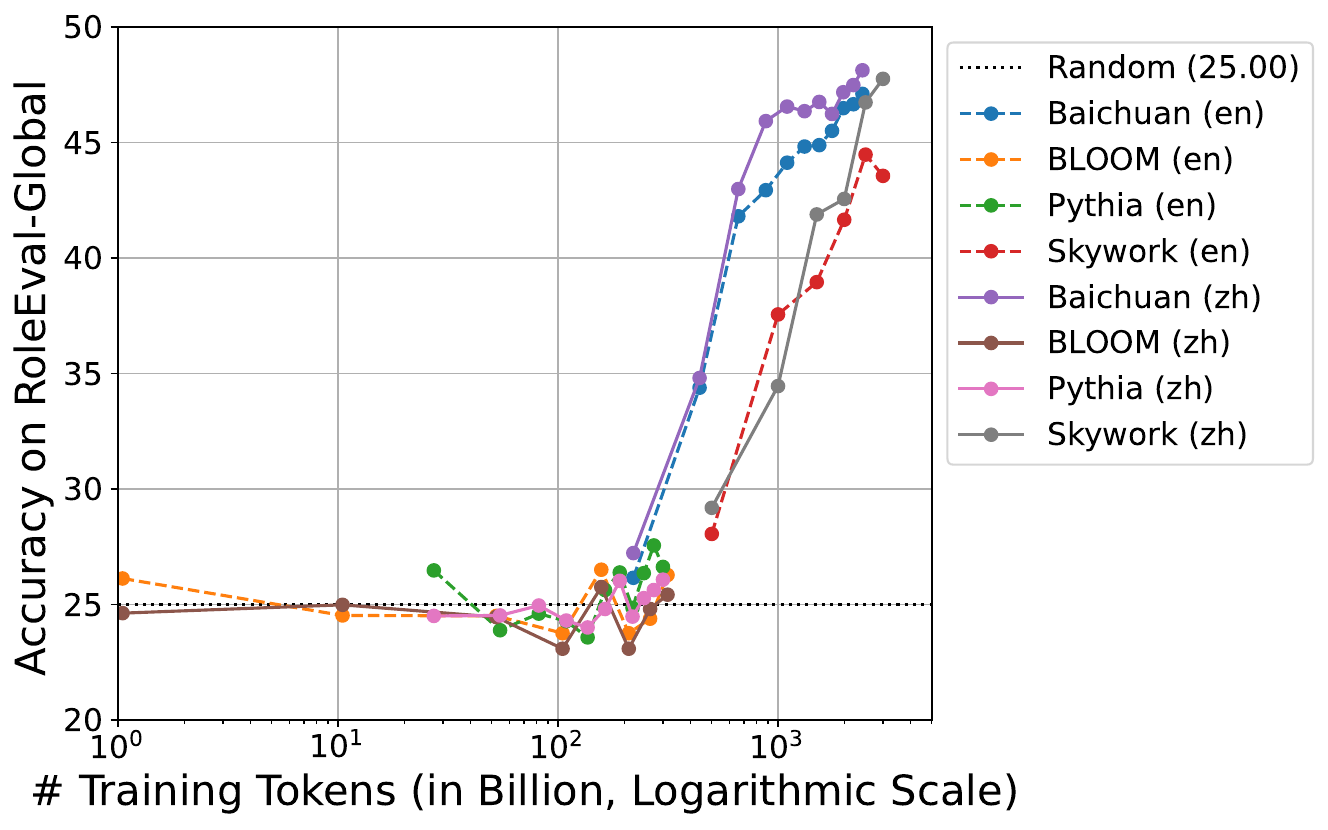}
    \caption{Token Scaling Laws on \textit{RoleEval-Global}.}
    \label{fig:token_scaling}
    
\end{figure}

Results from Figure \ref{fig:token_scaling} indicate that while the performance of Baichuan and Skywork with their first checkpoint (trained with 220B and 500B respectively) is near random, which is similar to BLOOM and Pythia, their subsequent checkpoints show steady improvement after 500B tokens, which indicates the importance of sufficient training tokens with fixed model size. However, the bottleneck of cross-lingual knowledge transfer can still be observed with increasing training tokens, which means that simply increasing the number of parameters and adding training tokens may not be the best way to break down the barriers between languages. In future research, we intend to examine other LLMs with intermediate checkpoints trained on larger-scale datasets primarily in English, aiming to substantiate our hypotheses with more robust evidence.

\section{Conclusion}

In this paper, we have presented RoleEval, a large-scale bilingual role evaluation benchmark, featuring 6,000 Chinese-English parallel questions across 300 diverse characters (200 for \textit{RoleEval-Global} and 100 for \textit{RoleEval-Chinese}) from five different domains. RoleEval is specifically designed to scrutinize the capabilities of LLMs in capturing, understanding, and reasoning over role knowledge. Our hybrid quality check process, blending automatic and human verification, guarantees the questions' difficulty and discrimination level, setting a new standard in benchmark design. Extensive evaluations of RoleEval on various LLMs, including both zero- and few-shot settings, highlight significant differences in knowledge distribution, as evidenced by GPT-4's superior performance in \textit{RoleEval-Global} and the notable excellence of Chinese LLMs in \textit{RoleEval-Chinese}. These findings not only demonstrate the disparities in LLMs' knowledge proficiency but also illuminate the path for future enhancements in bilingual and culture-specific LLMs. Through RoleEval, we aim to provide a robust framework for future advancements in LLM evaluation, particularly in role-playing scenarios, thereby enriching the landscape of language understanding and reasoning benchmarks.

\section*{Limitations}
In evaluating the effectiveness of real-world role evaluation benchmarks, there are two existing limitations. Firstly, the aspect of timeliness is crucial; the knowledge regarding real-world characters may change over time, making the benchmark outdated or irrelevant. To address this, we plan to explore methods for the automatic updating of benchmarks, ensuring that they remain current and reflective of ongoing changes. Secondly, the current format of benchmarks often restricts questions to having only one correct answer. This approach fails to adequately test scenarios where multiple answers could be correct, thus limiting the benchmark's ability to evaluate complex decision-making skills. A potential solution could be to incorporate a more dynamic question format that allows for the identification and acceptance of multiple correct answers, thereby enriching the assessment process by acknowledging the multi-faceted nature of real-world problems.

\section*{Ethics Statement}
Our benchmark is designed to enhance the model's understanding of role knowledge, which is crucial for improving persona consistency and factual accuracy while reducing hallucination. To achieve this, we have selected encyclopedic content that has been edited by multiple individuals. This approach helps to minimize factual errors and biases, particularly in comparison to other texts sourced from the Internet. Furthermore, all data collected for this project originates from publicly available materials, ensuring no concerns regarding privacy infringement. We confirm that all materials employed were utilized for non-commercial purposes, in adherence to copyright regulations and privacy policies.

Moreover, our goal is to promote a comprehensive and accurate understanding of role knowledge. Therefore, although most of our questions and options are positive, we have not entirely excluded potential negative aspects of the selected characters from our benchmark. Users should be aware of this when using this benchmark. Our commitment is to provide a comprehensive and balanced understanding, but users should remain critical and mindful of the context in which this information is used.

\bibliography{ref}

\begin{thebibliography}{34}
\expandafter\ifx\csname natexlab\endcsname\relax\def\natexlab#1{#1}\fi

\bibitem[{Ahn et~al.(2023)Ahn, Song, Yun, and Kim}]{ahnMPCHATMultimodalPersonaGrounded2023}
Jaewoo Ahn, Yeda Song, Sangdoo Yun, and Gunhee Kim. 2023.
\newblock \href {https://doi.org/10.18653/v1/2023.acl-long.189} {{{MPCHAT}}: {{Towards Multimodal Persona-Grounded Conversation}}}.
\newblock In \emph{Proceedings of the 61st {{Annual Meeting}} of the {{Association}} for {{Computational Linguistics}} ({{Volume}} 1: {{Long Papers}})}, pages 3354--3377, {Toronto, Canada}. {Association for Computational Linguistics}.

\bibitem[{Almazrouei et~al.(2023)Almazrouei, Alobeidli, Alshamsi, Cappelli, Cojocaru, Debbah, Goffinet, Hesslow, Launay, Malartic et~al.}]{almazrouei2023falcon}
Ebtesam Almazrouei, Hamza Alobeidli, Abdulaziz Alshamsi, Alessandro Cappelli, Ruxandra Cojocaru, M{\'e}rouane Debbah, {\'E}tienne Goffinet, Daniel Hesslow, Julien Launay, Quentin Malartic, et~al. 2023.
\newblock The falcon series of open language models.
\newblock \emph{arXiv preprint arXiv:2311.16867}.

\bibitem[{Bai et~al.(2023)Bai, Bai, Chu, Cui, Dang, Deng, Fan, Ge, Han, Huang, Hui, Ji, Li, Lin, Lin, Liu, Liu, Lu, Lu, Ma, Men, Ren, Ren, Tan, Tan, Tu, Wang, Wang, Wang, Wu, Xu, Xu, Yang, Yang, Yang, Yang, Yao, Yu, Yuan, Yuan, Zhang, Zhang, Zhang, Zhang, Zhou, Zhou, Zhou, and Zhu}]{baiQwenTechnicalReport2023}
Jinze Bai, Shuai Bai, Yunfei Chu, Zeyu Cui, Kai Dang, Xiaodong Deng, Yang Fan, Wenbin Ge, Yu~Han, Fei Huang, Binyuan Hui, Luo Ji, Mei Li, Junyang Lin, Runji Lin, Dayiheng Liu, Gao Liu, Chengqiang Lu, Keming Lu, Jianxin Ma, Rui Men, Xingzhang Ren, Xuancheng Ren, Chuanqi Tan, Sinan Tan, Jianhong Tu, Peng Wang, Shijie Wang, Wei Wang, Shengguang Wu, Benfeng Xu, Jin Xu, An~Yang, Hao Yang, Jian Yang, Shusheng Yang, Yang Yao, Bowen Yu, Hongyi Yuan, Zheng Yuan, Jianwei Zhang, Xingxuan Zhang, Yichang Zhang, Zhenru Zhang, Chang Zhou, Jingren Zhou, Xiaohuan Zhou, and Tianhang Zhu. 2023.
\newblock \href {https://arxiv.org/abs/2309.16609} {Qwen {{Technical Report}}}.
\newblock \emph{ArXiv preprint}, abs/2309.16609.

\bibitem[{Biderman et~al.(2023)Biderman, Schoelkopf, Anthony, Bradley, O'Brien, Hallahan, Khan, Purohit, Prashanth, Raff, Skowron, Sutawika, and {van der Wal}}]{bidermanPythiaSuiteAnalyzing2023}
Stella Biderman, Hailey Schoelkopf, Quentin Anthony, Herbie Bradley, Kyle O'Brien, Eric Hallahan, Mohammad~Aflah Khan, Shivanshu Purohit, USVSN~Sai Prashanth, Edward Raff, Aviya Skowron, Lintang Sutawika, and Oskar {van der Wal}. 2023.
\newblock \href {https://arxiv.org/abs/2304.01373} {Pythia: {{A Suite}} for {{Analyzing Large Language Models Across Training}} and {{Scaling}}}.
\newblock \emph{ArXiv preprint}, abs/2304.01373.

\bibitem[{Cui et~al.(2023)Cui, Yang, and Yao}]{cuiEfficientEffectiveText2023a}
Yiming Cui, Ziqing Yang, and Xin Yao. 2023.
\newblock \href {https://arxiv.org/abs/2304.08177} {Efficient and {{Effective Text Encoding}} for {{Chinese LLaMA}} and {{Alpaca}}}.
\newblock \emph{ArXiv preprint}, abs/2304.08177.

\bibitem[{Du et~al.(2022)Du, Qian, Liu, Ding, Qiu, Yang, and Tang}]{duGLMGeneralLanguage2022}
Zhengxiao Du, Yujie Qian, Xiao Liu, Ming Ding, Jiezhong Qiu, Zhilin Yang, and Jie Tang. 2022.
\newblock \href {https://doi.org/10.18653/v1/2022.acl-long.26} {{GLM}: General language model pretraining with autoregressive blank infilling}.
\newblock In \emph{Proceedings of the 60th Annual Meeting of the Association for Computational Linguistics (Volume 1: Long Papers)}, pages 320--335, Dublin, Ireland. Association for Computational Linguistics.

\bibitem[{Gao et~al.(2023)Gao, Tow, Abbasi, Biderman, Black, DiPofi, Foster, Golding, Hsu, Le~Noac'h, Li, McDonell, Muennighoff, Ociepa, Phang, Reynolds, Schoelkopf, Skowron, Sutawika, Tang, Thite, Wang, Wang, and Zou}]{gaoFrameworkFewshotLanguage2023}
Leo Gao, Jonathan Tow, Baber Abbasi, Stella Biderman, Sid Black, Anthony DiPofi, Charles Foster, Laurence Golding, Jeffrey Hsu, Alain Le~Noac'h, Haonan Li, Kyle McDonell, Niklas Muennighoff, Chris Ociepa, Jason Phang, Laria Reynolds, Hailey Schoelkopf, Aviya Skowron, Lintang Sutawika, Eric Tang, Anish Thite, Ben Wang, Kevin Wang, and Andy Zou. 2023.
\newblock \href {https://doi.org/10.5281/zenodo.10256836} {A framework for few-shot language model evaluation}.

\bibitem[{Hendrycks et~al.(2021)Hendrycks, Burns, Basart, Zou, Mazeika, Song, and Steinhardt}]{hendrycksMeasuringMassiveMultitask2021}
Dan Hendrycks, Collin Burns, Steven Basart, Andy Zou, Mantas Mazeika, Dawn Song, and Jacob Steinhardt. 2021.
\newblock \href {https://openreview.net/forum?id=d7KBjmI3GmQ} {Measuring massive multitask language understanding}.
\newblock In \emph{9th International Conference on Learning Representations, {ICLR} 2021, Virtual Event, Austria, May 3-7, 2021}. OpenReview.net.

\bibitem[{Huang et~al.(2023)Huang, Bai, Zhu, Zhang, Zhang, Su, Liu, Lv, Zhang, Lei, Fu, Sun, and He}]{huangCEvalMultiLevelMultiDiscipline2023}
Yuzhen Huang, Yuzhuo Bai, Zhihao Zhu, Junlei Zhang, Jinghan Zhang, Tangjun Su, Junteng Liu, Chuancheng Lv, Yikai Zhang, Jiayi Lei, Yao Fu, Maosong Sun, and Junxian He. 2023.
\newblock \href {https://arxiv.org/abs/2305.08322} {C-{{Eval}}: {{A Multi-Level Multi-Discipline Chinese Evaluation Suite}} for {{Foundation Models}}}.
\newblock \emph{ArXiv preprint}, abs/2305.08322.

\bibitem[{Jiang et~al.(2023)Jiang, Sablayrolles, Mensch, Bamford, Chaplot, de~las Casas, Bressand, Lengyel, Lample, Saulnier, Lavaud, Lachaux, Stock, Scao, Lavril, Wang, Lacroix, and Sayed}]{jiangMistral7B2023}
Albert~Q. Jiang, Alexandre Sablayrolles, Arthur Mensch, Chris Bamford, Devendra~Singh Chaplot, Diego de~las Casas, Florian Bressand, Gianna Lengyel, Guillaume Lample, Lucile Saulnier, L{\'e}lio~Renard Lavaud, Marie-Anne Lachaux, Pierre Stock, Teven~Le Scao, Thibaut Lavril, Thomas Wang, Timoth{\'e}e Lacroix, and William~El Sayed. 2023.
\newblock \href {https://arxiv.org/abs/2310.06825} {Mistral {{7B}}}.
\newblock \emph{ArXiv preprint}, abs/2310.06825.

\bibitem[{Li et~al.(2023{\natexlab{a}})Li, Leng, Yan, Shen, Wang, MI, Fei, Feng, Yan, Wang, Zhan, Jia, Wu, and Sun}]{liChatHaruhiRevivingAnime2023}
Cheng Li, Ziang Leng, Chenxi Yan, Junyi Shen, Hao Wang, Weishi MI, Yaying Fei, Xiaoyang Feng, Song Yan, HaoSheng Wang, Linkang Zhan, Yaokai Jia, Pingyu Wu, and Haozhen Sun. 2023{\natexlab{a}}.
\newblock \href {https://arxiv.org/abs/2308.09597} {{{ChatHaruhi}}: {{Reviving Anime Character}} in {{Reality}} via {{Large Language Model}}}.
\newblock \emph{ArXiv preprint}, abs/2308.09597.

\bibitem[{Li et~al.(2023{\natexlab{b}})Li, Zhang, Koto, Yang, Zhao, Gong, Duan, and Baldwin}]{liCMMLUMeasuringMassive2023}
Haonan Li, Yixuan Zhang, Fajri Koto, Yifei Yang, Hai Zhao, Yeyun Gong, Nan Duan, and Timothy Baldwin. 2023{\natexlab{b}}.
\newblock \href {https://arxiv.org/abs/2306.09212} {{{CMMLU}}: {{Measuring}} massive multitask language understanding in {{Chinese}}}.
\newblock \emph{ArXiv preprint}, abs/2306.09212.

\bibitem[{Lin et~al.(2023)Lin, Ravichander, Lu, Dziri, Sclar, Chandu, Bhagavatula, and Choi}]{linUnlockingSpellBase2023}
Bill~Yuchen Lin, Abhilasha Ravichander, Ximing Lu, Nouha Dziri, Melanie Sclar, Khyathi Chandu, Chandra Bhagavatula, and Yejin Choi. 2023.
\newblock \href {https://arxiv.org/abs/2312.01552} {The {{Unlocking Spell}} on {{Base LLMs}}: {{Rethinking Alignment}} via {{In-Context Learning}}}.
\newblock \emph{ArXiv preprint}, abs/2312.01552.

\bibitem[{Liu et~al.(2021)Liu, Zheng, Demasi, Sabour, Li, Yu, Jiang, and Huang}]{liuEmotionalSupportDialog2021}
Siyang Liu, Chujie Zheng, Orianna Demasi, Sahand Sabour, Yu~Li, Zhou Yu, Yong Jiang, and Minlie Huang. 2021.
\newblock \href {https://doi.org/10.18653/v1/2021.acl-long.269} {Towards emotional support dialog systems}.
\newblock In \emph{Proceedings of the 59th Annual Meeting of the Association for Computational Linguistics and the 11th International Joint Conference on Natural Language Processing (Volume 1: Long Papers)}, pages 3469--3483, Online. Association for Computational Linguistics.

\bibitem[{OpenAI(2023)}]{openaiGPT4TechnicalReport2023}
OpenAI. 2023.
\newblock \href {https://arxiv.org/abs/2303.08774} {{{GPT-4 Technical Report}}}.
\newblock \emph{ArXiv preprint}, abs/2303.08774.

\bibitem[{Ouyang et~al.(2022)Ouyang, Wu, Jiang, Almeida, Wainwright, Mishkin, Zhang, Agarwal, Slama, Ray, Schulman, Hilton, Kelton, Miller, Simens, Askell, Welinder, Christiano, Leike, and Lowe}]{ouyangTrainingLanguageModels2022}
Long Ouyang, Jeff Wu, Xu~Jiang, Diogo Almeida, Carroll~L. Wainwright, Pamela Mishkin, Chong Zhang, Sandhini Agarwal, Katarina Slama, Alex Ray, John Schulman, Jacob Hilton, Fraser Kelton, Luke Miller, Maddie Simens, Amanda Askell, Peter Welinder, Paul Christiano, Jan Leike, and Ryan Lowe. 2022.
\newblock \href {https://arxiv.org/abs/2203.02155} {Training language models to follow instructions with human feedback}.
\newblock \emph{ArXiv preprint}, abs/2203.02155.

\bibitem[{Shao et~al.(2023)Shao, Li, Dai, and Qiu}]{shaoCharacterLLMTrainableAgent2023}
Yunfan Shao, Linyang Li, Junqi Dai, and Xipeng Qiu. 2023.
\newblock \href {https://arxiv.org/abs/2310.10158} {Character-{{LLM}}: {{A Trainable Agent}} for {{Role-Playing}}}.
\newblock \emph{ArXiv preprint}, abs/2310.10158.

\bibitem[{Shuster et~al.(2022)Shuster, Urbanek, Szlam, and Weston}]{shusterAmMeYou2021}
Kurt Shuster, Jack Urbanek, Arthur Szlam, and Jason Weston. 2022.
\newblock \href {https://doi.org/10.18653/v1/2022.findings-naacl.182} {Am {I} me or you? state-of-the-art dialogue models cannot maintain an identity}.
\newblock In \emph{Findings of the Association for Computational Linguistics: NAACL 2022}, pages 2367--2387, Seattle, United States. Association for Computational Linguistics.

\bibitem[{Touvron et~al.(2023{\natexlab{a}})Touvron, Lavril, Izacard, Martinet, Lachaux, Lacroix, Rozi{\`e}re, Goyal, Hambro, Azhar, Rodriguez, Joulin, Grave, and Lample}]{touvronLLaMAOpenEfficient2023}
Hugo Touvron, Thibaut Lavril, Gautier Izacard, Xavier Martinet, Marie-Anne Lachaux, Timoth{\'e}e Lacroix, Baptiste Rozi{\`e}re, Naman Goyal, Eric Hambro, Faisal Azhar, Aurelien Rodriguez, Armand Joulin, Edouard Grave, and Guillaume Lample. 2023{\natexlab{a}}.
\newblock \href {https://arxiv.org/abs/2302.13971} {{{LLaMA}}: {{Open}} and {{Efficient Foundation Language Models}}}.
\newblock \emph{ArXiv preprint}, abs/2302.13971.

\bibitem[{Touvron et~al.(2023{\natexlab{b}})Touvron, Martin, Stone, Albert, Almahairi, Babaei, Bashlykov, Batra, Bhargava, Bhosale, Bikel, Blecher, Ferrer, Chen, Cucurull, Esiobu, Fernandes, Fu, Fu, Fuller, Gao, Goswami, Goyal, Hartshorn, Hosseini, Hou, Inan, Kardas, Kerkez, Khabsa, Kloumann, Korenev, Koura, Lachaux, Lavril, Lee, Liskovich, Lu, Mao, Martinet, Mihaylov, Mishra, Molybog, Nie, Poulton, Reizenstein, Rungta, Saladi, Schelten, Silva, Smith, Subramanian, Tan, Tang, Taylor, Williams, Kuan, Xu, Yan, Zarov, Zhang, Fan, Kambadur, Narang, Rodriguez, Stojnic, Edunov, and Scialom}]{touvronLlamaOpenFoundation2023}
Hugo Touvron, Louis Martin, Kevin Stone, Peter Albert, Amjad Almahairi, Yasmine Babaei, Nikolay Bashlykov, Soumya Batra, Prajjwal Bhargava, Shruti Bhosale, Dan Bikel, Lukas Blecher, Cristian~Canton Ferrer, Moya Chen, Guillem Cucurull, David Esiobu, Jude Fernandes, Jeremy Fu, Wenyin Fu, Brian Fuller, Cynthia Gao, Vedanuj Goswami, Naman Goyal, Anthony Hartshorn, Saghar Hosseini, Rui Hou, Hakan Inan, Marcin Kardas, Viktor Kerkez, Madian Khabsa, Isabel Kloumann, Artem Korenev, Punit~Singh Koura, Marie-Anne Lachaux, Thibaut Lavril, Jenya Lee, Diana Liskovich, Yinghai Lu, Yuning Mao, Xavier Martinet, Todor Mihaylov, Pushkar Mishra, Igor Molybog, Yixin Nie, Andrew Poulton, Jeremy Reizenstein, Rashi Rungta, Kalyan Saladi, Alan Schelten, Ruan Silva, Eric~Michael Smith, Ranjan Subramanian, Xiaoqing~Ellen Tan, Binh Tang, Ross Taylor, Adina Williams, Jian~Xiang Kuan, Puxin Xu, Zheng Yan, Iliyan Zarov, Yuchen Zhang, Angela Fan, Melanie Kambadur, Sharan Narang, Aurelien Rodriguez, Robert Stojnic, Sergey Edunov, and Thomas
  Scialom. 2023{\natexlab{b}}.
\newblock \href {https://arxiv.org/abs/2307.09288} {Llama 2: {{Open Foundation}} and {{Fine-Tuned Chat Models}}}.
\newblock \emph{ArXiv preprint}, abs/2307.09288.

\bibitem[{Wang et~al.(2023{\natexlab{a}})Wang, Ma, Feng, Zhang, Yang, Zhang, Chen, Tang, Chen, Lin, Zhao, Wei, and Wen}]{wangSurveyLargeLanguage2023}
Lei Wang, Chen Ma, Xueyang Feng, Zeyu Zhang, Hao Yang, Jingsen Zhang, Zhiyuan Chen, Jiakai Tang, Xu~Chen, Yankai Lin, Wayne~Xin Zhao, Zhewei Wei, and Ji-Rong Wen. 2023{\natexlab{a}}.
\newblock \href {https://arxiv.org/abs/2308.11432} {A {{Survey}} on {{Large Language Model}} based {{Autonomous Agents}}}.
\newblock \emph{ArXiv preprint}, abs/2308.11432.

\bibitem[{Wang et~al.(2023{\natexlab{b}})Wang, Peng, Que, Liu, Zhou, Wu, Guo, Gan, Ni, Zhang, Zhang, Ouyang, Xu, Chen, Fu, and Peng}]{wangRoleLLMBenchmarkingEliciting2023}
Zekun~Moore Wang, Zhongyuan Peng, Haoran Que, Jiaheng Liu, Wangchunshu Zhou, Yuhan Wu, Hongcheng Guo, Ruitong Gan, Zehao Ni, Man Zhang, Zhaoxiang Zhang, Wanli Ouyang, Ke~Xu, Wenhu Chen, Jie Fu, and Junran Peng. 2023{\natexlab{b}}.
\newblock \href {https://arxiv.org/abs/2310.00746} {{{RoleLLM}}: {{Benchmarking}}, {{Eliciting}}, and {{Enhancing Role-Playing Abilities}} of {{Large Language Models}}}.
\newblock \emph{ArXiv preprint}, abs/2310.00746.

\bibitem[{Wei et~al.(2023{\natexlab{a}})Wei, Shuster, Szlam, Weston, Urbanek, and Komeili}]{weiMultiPartyChatConversational2023}
Jimmy Wei, Kurt Shuster, Arthur Szlam, Jason Weston, Jack Urbanek, and Mojtaba Komeili. 2023{\natexlab{a}}.
\newblock \href {https://arxiv.org/abs/2304.13835} {Multi-{{Party Chat}}: {{Conversational Agents}} in {{Group Settings}} with {{Humans}} and {{Models}}}.
\newblock \emph{ArXiv preprint}, abs/2304.13835.

\bibitem[{Wei et~al.(2023{\natexlab{b}})Wei, Zhao, Zhang, Zhu, Wang, Yang, Li, Cheng, L{\"u}, Hu, Li, Yang, Luo, Wu, Liu, Cheng, Cheng, Zhang, Zhang, Lin, Wang, Ma, Dong, Sun, Chen, Peng, Liang, Yan, Fang, and Zhou}]{weiSkyworkMoreOpen2023}
Tianwen Wei, Liang Zhao, Lichang Zhang, Bo~Zhu, Lijie Wang, Haihua Yang, Biye Li, Cheng Cheng, Weiwei L{\"u}, Rui Hu, Chenxia Li, Liu Yang, Xilin Luo, Xuejie Wu, Lunan Liu, Wenjun Cheng, Peng Cheng, Jianhao Zhang, Xiaoyu Zhang, Lei Lin, Xiaokun Wang, Yutuan Ma, Chuanhai Dong, Yanqi Sun, Yifu Chen, Yongyi Peng, Xiaojuan Liang, Shuicheng Yan, Han Fang, and Yahui Zhou. 2023{\natexlab{b}}.
\newblock \href {https://arxiv.org/abs/2310.19341} {Skywork: {{A More Open Bilingual Foundation Model}}}.
\newblock \emph{ArXiv preprint}, abs/2310.19341.

\bibitem[{Weizenbaum(1966)}]{weizenbaumELIZAComputerProgram1966}
Joseph Weizenbaum. 1966.
\newblock \href {https://doi.org/10.1145/365153.365168} {{{ELIZA}}{\textemdash}a computer program for the study of natural language communication between man and machine}.
\newblock \emph{Communications of the ACM}, 9(1):36--45.

\bibitem[{Workshop et~al.(2022)Workshop, Scao, Fan, Akiki, Pavlick, Ili{\'c}, Hesslow, Castagn{\'e}, Luccioni, Yvon, Gall{\'e}, Tow, Rush, Biderman, Webson, Ammanamanchi, Wang, Sagot, Muennighoff, {del Moral}, Ruwase, Bawden, Bekman, {McMillan-Major}, Beltagy, Nguyen, Saulnier, Tan, Suarez, Sanh, Lauren{\c c}on, Jernite, Launay, Mitchell, Raffel, Gokaslan, Simhi, Soroa, Aji, Alfassy, Rogers, Nitzav, Xu, Mou, Emezue, Klamm, Leong, {van Strien}, Adelani, Radev, Ponferrada, Levkovizh, Kim, Natan, De~Toni, Dupont, Kruszewski, Pistilli, Elsahar, Benyamina, Tran, Yu, Abdulmumin, Johnson, {Gonzalez-Dios}, {de la Rosa}, Chim, Dodge, Zhu, Chang, Frohberg, Tobing, Bhattacharjee, Almubarak, Chen, Lo, Von~Werra, Weber, Phan, {allal}, Tanguy, Dey, Mu{\~n}oz, Masoud, Grandury, {\v S}a{\v s}ko, Huang, Coavoux, Singh, Jiang, Vu, Jauhar, Ghaleb, Subramani, Kassner, Khamis, Nguyen, Espejel, {de Gibert}, Villegas, Henderson, Colombo, Amuok, Lhoest, Harliman, Bommasani, L{\'o}pez, Ribeiro, Osei, Pyysalo, Nagel, Bose, Muhammad,
  Sharma, Longpre, Nikpoor, Silberberg, Pai, Zink, Torrent, Schick, Thrush, Danchev, Nikoulina, Laippala, Lepercq, Prabhu, Alyafeai, Talat, Raja, Heinzerling, Si, Ta{\c s}ar, Salesky, Mielke, Lee, Sharma, Santilli, Chaffin, Stiegler, Datta, Szczechla, Chhablani, Wang, Pandey, Strobelt, Fries, Rozen, Gao, Sutawika, Bari, {Al-shaibani}, Manica, Nayak, Teehan, Albanie, Shen, {Ben-David}, Bach, Kim, Bers, Fevry, Neeraj, Thakker, Raunak, Tang, Yong, Sun, Brody, Uri, Tojarieh, Roberts, Chung, Tae, Phang, Press, Li, Narayanan, Bourfoune, Casper, Rasley, Ryabinin, Mishra, Zhang, Shoeybi, Peyrounette, Patry, Tazi, Sanseviero, {von Platen}, Cornette, Lavall{\'e}e, Lacroix, Rajbhandari, Gandhi, Smith, Requena, Patil, Dettmers, Baruwa, Singh, Cheveleva, Ligozat, Subramonian, N{\'e}v{\'e}ol, Lovering, Garrette, Tunuguntla, Reiter, Taktasheva, Voloshina, Bogdanov, Winata, Schoelkopf, Kalo, Novikova, Forde, Clive, Kasai, Kawamura, Hazan, Carpuat, Clinciu, Kim, Cheng, Serikov, Antverg, {van der Wal}, Zhang, Zhang, Gehrmann,
  Mirkin, Pais, Shavrina, Scialom, Yun, Limisiewicz, Rieser, Protasov, Mikhailov, Pruksachatkun, Belinkov, Bamberger, Kasner, Rueda, Pestana, Feizpour, Khan, Faranak, Santos, Hevia, Unldreaj, Aghagol, Abdollahi, Tammour, HajiHosseini, Behroozi, Ajibade, Saxena, Ferrandis, McDuff, Contractor, Lansky, David, Kiela, Nguyen, Tan, Baylor, Ozoani, Mirza, Ononiwu, Rezanejad, Jones, Bhattacharya, Solaiman, Sedenko, Nejadgholi, Passmore, Seltzer, Sanz, Dutra, Samagaio, Elbadri, Mieskes, Gerchick, Akinlolu, McKenna, Qiu, Ghauri, Burynok, Abrar, Rajani, Elkott, Fahmy, Samuel, An, Kromann, Hao, Alizadeh, Shubber, Wang, Roy, Viguier, Le, Oyebade, Le, Yang, Nguyen, Kashyap, Palasciano, Callahan, Shukla, {Miranda-Escalada}, Singh, Beilharz, Wang, Brito, Zhou, Jain, Xu, Fourrier, Peri{\~n}{\'a}n, Molano, Yu, Manjavacas, Barth, Fuhrimann, Altay, Bayrak, Burns, Vrabec, Bello, Dash, Kang, Giorgi, Golde, Posada, Sivaraman, Bulchandani, Liu, Shinzato, {de Bykhovetz}, Takeuchi, P{\`a}mies, Castillo, Nezhurina, S{\"a}nger, Samwald,
  Cullan, Weinberg, De~Wolf, Mihaljcic, Liu, Freidank, Kang, Seelam, Dahlberg, Broad, Muellner, Fung, Haller, Chandrasekhar, Eisenberg, Martin, Canalli, Su, Su, Cahyawijaya, Garda, Deshmukh, Mishra, Kiblawi, Ott, {Sang-aroonsiri}, Kumar, Schweter, Bharati, Laud, Gigant, Kainuma, Kusa, Labrak, Bajaj, Venkatraman, Xu, Xu, Xu, Tan, Xie, Ye, Bras, Belkada, and Wolf}]{workshopBLOOM176BParameterOpenAccess2023}
BigScience Workshop, Teven~Le Scao, Angela Fan, Christopher Akiki, Ellie Pavlick, Suzana Ili{\'c}, Daniel Hesslow, Roman Castagn{\'e}, Alexandra~Sasha Luccioni, Fran{\c c}ois Yvon, Matthias Gall{\'e}, Jonathan Tow, Alexander~M. Rush, Stella Biderman, Albert Webson, Pawan~Sasanka Ammanamanchi, Thomas Wang, Beno{\^i}t Sagot, Niklas Muennighoff, Albert~Villanova {del Moral}, Olatunji Ruwase, Rachel Bawden, Stas Bekman, Angelina {McMillan-Major}, Iz~Beltagy, Huu Nguyen, Lucile Saulnier, Samson Tan, Pedro~Ortiz Suarez, Victor Sanh, Hugo Lauren{\c c}on, Yacine Jernite, Julien Launay, Margaret Mitchell, Colin Raffel, Aaron Gokaslan, Adi Simhi, Aitor Soroa, Alham~Fikri Aji, Amit Alfassy, Anna Rogers, Ariel~Kreisberg Nitzav, Canwen Xu, Chenghao Mou, Chris Emezue, Christopher Klamm, Colin Leong, Daniel {van Strien}, David~Ifeoluwa Adelani, Dragomir Radev, Eduardo~Gonz{\'a}lez Ponferrada, Efrat Levkovizh, Ethan Kim, Eyal~Bar Natan, Francesco De~Toni, G{\'e}rard Dupont, Germ{\'a}n Kruszewski, Giada Pistilli, Hady
  Elsahar, Hamza Benyamina, Hieu Tran, Ian Yu, Idris Abdulmumin, Isaac Johnson, Itziar {Gonzalez-Dios}, Javier {de la Rosa}, Jenny Chim, Jesse Dodge, Jian Zhu, Jonathan Chang, J{\"o}rg Frohberg, Joseph Tobing, Joydeep Bhattacharjee, Khalid Almubarak, Kimbo Chen, Kyle Lo, Leandro Von~Werra, Leon Weber, Long Phan, Loubna~Ben {allal}, Ludovic Tanguy, Manan Dey, Manuel~Romero Mu{\~n}oz, Maraim Masoud, Mar{\'i}a Grandury, Mario {\v S}a{\v s}ko, Max Huang, Maximin Coavoux, Mayank Singh, Mike Tian-Jian Jiang, Minh~Chien Vu, Mohammad~A. Jauhar, Mustafa Ghaleb, Nishant Subramani, Nora Kassner, Nurulaqilla Khamis, Olivier Nguyen, Omar Espejel, Ona {de Gibert}, Paulo Villegas, Peter Henderson, Pierre Colombo, Priscilla Amuok, Quentin Lhoest, Rheza Harliman, Rishi Bommasani, Roberto~Luis L{\'o}pez, Rui Ribeiro, Salomey Osei, Sampo Pyysalo, Sebastian Nagel, Shamik Bose, Shamsuddeen~Hassan Muhammad, Shanya Sharma, Shayne Longpre, Somaieh Nikpoor, Stanislav Silberberg, Suhas Pai, Sydney Zink, Tiago~Timponi Torrent, Timo
  Schick, Tristan Thrush, Valentin Danchev, Vassilina Nikoulina, Veronika Laippala, Violette Lepercq, Vrinda Prabhu, Zaid Alyafeai, Zeerak Talat, Arun Raja, Benjamin Heinzerling, Chenglei Si, Davut~Emre Ta{\c s}ar, Elizabeth Salesky, Sabrina~J. Mielke, Wilson~Y. Lee, Abheesht Sharma, Andrea Santilli, Antoine Chaffin, Arnaud Stiegler, Debajyoti Datta, Eliza Szczechla, Gunjan Chhablani, Han Wang, Harshit Pandey, Hendrik Strobelt, Jason~Alan Fries, Jos Rozen, Leo Gao, Lintang Sutawika, M.~Saiful Bari, Maged~S. {Al-shaibani}, Matteo Manica, Nihal Nayak, Ryan Teehan, Samuel Albanie, Sheng Shen, Srulik {Ben-David}, Stephen~H. Bach, Taewoon Kim, Tali Bers, Thibault Fevry, Trishala Neeraj, Urmish Thakker, Vikas Raunak, Xiangru Tang, Zheng-Xin Yong, Zhiqing Sun, Shaked Brody, Yallow Uri, Hadar Tojarieh, Adam Roberts, Hyung~Won Chung, Jaesung Tae, Jason Phang, Ofir Press, Conglong Li, Deepak Narayanan, Hatim Bourfoune, Jared Casper, Jeff Rasley, Max Ryabinin, Mayank Mishra, Minjia Zhang, Mohammad Shoeybi, Myriam
  Peyrounette, Nicolas Patry, Nouamane Tazi, Omar Sanseviero, Patrick {von Platen}, Pierre Cornette, Pierre~Fran{\c c}ois Lavall{\'e}e, R{\'e}mi Lacroix, Samyam Rajbhandari, Sanchit Gandhi, Shaden Smith, St{\'e}phane Requena, Suraj Patil, Tim Dettmers, Ahmed Baruwa, Amanpreet Singh, Anastasia Cheveleva, Anne-Laure Ligozat, Arjun Subramonian, Aur{\'e}lie N{\'e}v{\'e}ol, Charles Lovering, Dan Garrette, Deepak Tunuguntla, Ehud Reiter, Ekaterina Taktasheva, Ekaterina Voloshina, Eli Bogdanov, Genta~Indra Winata, Hailey Schoelkopf, Jan-Christoph Kalo, Jekaterina Novikova, Jessica~Zosa Forde, Jordan Clive, Jungo Kasai, Ken Kawamura, Liam Hazan, Marine Carpuat, Miruna Clinciu, Najoung Kim, Newton Cheng, Oleg Serikov, Omer Antverg, Oskar {van der Wal}, Rui Zhang, Ruochen Zhang, Sebastian Gehrmann, Shachar Mirkin, Shani Pais, Tatiana Shavrina, Thomas Scialom, Tian Yun, Tomasz Limisiewicz, Verena Rieser, Vitaly Protasov, Vladislav Mikhailov, Yada Pruksachatkun, Yonatan Belinkov, Zachary Bamberger, Zden{\v e}k Kasner,
  Alice Rueda, Amanda Pestana, Amir Feizpour, Ammar Khan, Amy Faranak, Ana Santos, Anthony Hevia, Antigona Unldreaj, Arash Aghagol, Arezoo Abdollahi, Aycha Tammour, Azadeh HajiHosseini, Bahareh Behroozi, Benjamin Ajibade, Bharat Saxena, Carlos~Mu{\~n}oz Ferrandis, Daniel McDuff, Danish Contractor, David Lansky, Davis David, Douwe Kiela, Duong~A. Nguyen, Edward Tan, Emi Baylor, Ezinwanne Ozoani, Fatima Mirza, Frankline Ononiwu, Habib Rezanejad, Hessie Jones, Indrani Bhattacharya, Irene Solaiman, Irina Sedenko, Isar Nejadgholi, Jesse Passmore, Josh Seltzer, Julio~Bonis Sanz, Livia Dutra, Mairon Samagaio, Maraim Elbadri, Margot Mieskes, Marissa Gerchick, Martha Akinlolu, Michael McKenna, Mike Qiu, Muhammed Ghauri, Mykola Burynok, Nafis Abrar, Nazneen Rajani, Nour Elkott, Nour Fahmy, Olanrewaju Samuel, Ran An, Rasmus Kromann, Ryan Hao, Samira Alizadeh, Sarmad Shubber, Silas Wang, Sourav Roy, Sylvain Viguier, Thanh Le, Tobi Oyebade, Trieu Le, Yoyo Yang, Zach Nguyen, Abhinav~Ramesh Kashyap, Alfredo Palasciano,
  Alison Callahan, Anima Shukla, Antonio {Miranda-Escalada}, Ayush Singh, Benjamin Beilharz, Bo~Wang, Caio Brito, Chenxi Zhou, Chirag Jain, Chuxin Xu, Cl{\'e}mentine Fourrier, Daniel~Le{\'o}n Peri{\~n}{\'a}n, Daniel Molano, Dian Yu, Enrique Manjavacas, Fabio Barth, Florian Fuhrimann, Gabriel Altay, Giyaseddin Bayrak, Gully Burns, Helena~U. Vrabec, Imane Bello, Ishani Dash, Jihyun Kang, John Giorgi, Jonas Golde, Jose~David Posada, Karthik~Rangasai Sivaraman, Lokesh Bulchandani, Lu~Liu, Luisa Shinzato, Madeleine~Hahn {de Bykhovetz}, Maiko Takeuchi, Marc P{\`a}mies, Maria~A. Castillo, Marianna Nezhurina, Mario S{\"a}nger, Matthias Samwald, Michael Cullan, Michael Weinberg, Michiel De~Wolf, Mina Mihaljcic, Minna Liu, Moritz Freidank, Myungsun Kang, Natasha Seelam, Nathan Dahlberg, Nicholas~Michio Broad, Nikolaus Muellner, Pascale Fung, Patrick Haller, Ramya Chandrasekhar, Renata Eisenberg, Robert Martin, Rodrigo Canalli, Rosaline Su, Ruisi Su, Samuel Cahyawijaya, Samuele Garda, Shlok~S. Deshmukh, Shubhanshu
  Mishra, Sid Kiblawi, Simon Ott, Sinee {Sang-aroonsiri}, Srishti Kumar, Stefan Schweter, Sushil Bharati, Tanmay Laud, Th{\'e}o Gigant, Tomoya Kainuma, Wojciech Kusa, Yanis Labrak, Yash~Shailesh Bajaj, Yash Venkatraman, Yifan Xu, Yingxin Xu, Yu~Xu, Zhe Tan, Zhongli Xie, Zifan Ye, Mathilde Bras, Younes Belkada, and Thomas Wolf. 2022.
\newblock \href {https://arxiv.org/abs/2211.05100} {{{BLOOM}}: {{A 176B-Parameter Open-Access Multilingual Language Model}}}.
\newblock \emph{ArXiv preprint}, abs/2211.05100.

\bibitem[{Xi et~al.(2023)Xi, Chen, Guo, He, Ding, Hong, Zhang, Wang, Jin, Zhou, Zheng, Fan, Wang, Xiong, Zhou, Wang, Jiang, Zou, Liu, Yin, Dou, Weng, Cheng, Zhang, Qin, Zheng, Qiu, Huang, and Gui}]{xiRisePotentialLarge2023}
Zhiheng Xi, Wenxiang Chen, Xin Guo, Wei He, Yiwen Ding, Boyang Hong, Ming Zhang, Junzhe Wang, Senjie Jin, Enyu Zhou, Rui Zheng, Xiaoran Fan, Xiao Wang, Limao Xiong, Yuhao Zhou, Weiran Wang, Changhao Jiang, Yicheng Zou, Xiangyang Liu, Zhangyue Yin, Shihan Dou, Rongxiang Weng, Wensen Cheng, Qi~Zhang, Wenjuan Qin, Yongyan Zheng, Xipeng Qiu, Xuanjing Huang, and Tao Gui. 2023.
\newblock \href {https://arxiv.org/abs/2309.07864} {The {{Rise}} and {{Potential}} of {{Large Language Model Based Agents}}: {{A Survey}}}.
\newblock \emph{ArXiv preprint}, abs/2309.07864.

\bibitem[{Xu et~al.(2022)Xu, Gou, Wu, Niu, Wu, Wang, and Wang}]{xuLongTimeNo2022a}
Xinchao Xu, Zhibin Gou, Wenquan Wu, Zheng-Yu Niu, Hua Wu, Haifeng Wang, and Shihang Wang. 2022.
\newblock \href {https://doi.org/10.18653/v1/2022.findings-acl.207} {Long time no see! open-domain conversation with long-term persona memory}.
\newblock In \emph{Findings of the Association for Computational Linguistics: ACL 2022}, pages 2639--2650, Dublin, Ireland. Association for Computational Linguistics.

\bibitem[{Yang et~al.(2023)Yang, Xiao, Wang, Zhang, Bian, Yin, Lv, Pan, Wang, Yan, Yang, Deng, Wang, Liu, Ai, Dong, Zhao, Xu, Sun, Zhang, Liu, Ji, Xie, Dai, Fang, Su, Song, Liu, Ru, Ma, Wang, Liu, Lin, Nie, Guo, Sun, Zhang, Li, Li, Cheng, Chen, Zeng, Wang, Chen, Men, Yu, Pan, Shen, Wang, Li, Jiang, Gao, Zhang, Zhou, and Wu}]{yangBaichuanOpenLargescale2023}
Aiyuan Yang, Bin Xiao, Bingning Wang, Borong Zhang, Ce~Bian, Chao Yin, Chenxu Lv, Da~Pan, Dian Wang, Dong Yan, Fan Yang, Fei Deng, Feng Wang, Feng Liu, Guangwei Ai, Guosheng Dong, Haizhou Zhao, Hang Xu, Haoze Sun, Hongda Zhang, Hui Liu, Jiaming Ji, Jian Xie, JunTao Dai, Kun Fang, Lei Su, Liang Song, Lifeng Liu, Liyun Ru, Luyao Ma, Mang Wang, Mickel Liu, MingAn Lin, Nuolan Nie, Peidong Guo, Ruiyang Sun, Tao Zhang, Tianpeng Li, Tianyu Li, Wei Cheng, Weipeng Chen, Xiangrong Zeng, Xiaochuan Wang, Xiaoxi Chen, Xin Men, Xin Yu, Xuehai Pan, Yanjun Shen, Yiding Wang, Yiyu Li, Youxin Jiang, Yuchen Gao, Yupeng Zhang, Zenan Zhou, and Zhiying Wu. 2023.
\newblock \href {https://arxiv.org/abs/2309.10305} {Baichuan 2: {{Open Large-scale Language Models}}}.
\newblock \emph{ArXiv preprint}, abs/2309.10305.

\bibitem[{Zeng et~al.(2022)Zeng, Liu, Du, Wang, Lai, Ding, Yang, Xu, Zheng, Xia, Tam, Ma, Xue, Zhai, Chen, Zhang, Dong, and Tang}]{zengGLM130BOpenBilingual2023}
Aohan Zeng, Xiao Liu, Zhengxiao Du, Zihan Wang, Hanyu Lai, Ming Ding, Zhuoyi Yang, Yifan Xu, Wendi Zheng, Xiao Xia, Weng~Lam Tam, Zixuan Ma, Yufei Xue, Jidong Zhai, Wenguang Chen, Peng Zhang, Yuxiao Dong, and Jie Tang. 2022.
\newblock \href {https://arxiv.org/abs/2210.02414} {{{GLM-130B}}: {{An Open Bilingual Pre-trained Model}}}.
\newblock \emph{ArXiv preprint}, abs/2210.02414.

\bibitem[{Zhang et~al.(2018)Zhang, Dinan, Urbanek, Szlam, Kiela, and Weston}]{zhangPersonalizingDialogueAgents2018}
Saizheng Zhang, Emily Dinan, Jack Urbanek, Arthur Szlam, Douwe Kiela, and Jason Weston. 2018.
\newblock \href {https://doi.org/10.18653/v1/P18-1205} {Personalizing dialogue agents: {I} have a dog, do you have pets too?}
\newblock In \emph{Proceedings of the 56th Annual Meeting of the Association for Computational Linguistics (Volume 1: Long Papers)}, pages 2204--2213, Melbourne, Australia. Association for Computational Linguistics.

\bibitem[{Zheng et~al.(2019)Zheng, Chen, Huang, Liu, and Zhu}]{zhengPersonalizedDialogueGeneration2020}
Yinhe Zheng, Guanyi Chen, Minlie Huang, Song Liu, and Xuan Zhu. 2019.
\newblock \href {https://arxiv.org/abs/1901.09672} {Personalized {{Dialogue Generation}} with {{Diversified Traits}}}.
\newblock \emph{ArXiv preprint}, abs/1901.09672.

\bibitem[{Zhou et~al.(2023{\natexlab{a}})Zhou, Liu, Xu, Iyer, Sun, Mao, Ma, Efrat, Yu, Yu, Zhang, Ghosh, Lewis, Zettlemoyer, and Levy}]{zhouLIMALessMore2023a}
Chunting Zhou, Pengfei Liu, Puxin Xu, Srini Iyer, Jiao Sun, Yuning Mao, Xuezhe Ma, Avia Efrat, Ping Yu, Lili Yu, Susan Zhang, Gargi Ghosh, Mike Lewis, Luke Zettlemoyer, and Omer Levy. 2023{\natexlab{a}}.
\newblock \href {https://arxiv.org/abs/2305.11206} {{{LIMA}}: {{Less Is More}} for {{Alignment}}}.
\newblock \emph{ArXiv preprint}, abs/2305.11206.

\bibitem[{Zhou et~al.(2023{\natexlab{b}})Zhou, Chen, Wan, Wen, Song, Yu, Huang, Peng, Yang, Xiao, Sabour, Zhang, Hou, Zhang, Dong, Tang, and Huang}]{zhouCharacterGLMCustomizingChinese2023}
Jinfeng Zhou, Zhuang Chen, Dazhen Wan, Bosi Wen, Yi~Song, Jifan Yu, Yongkang Huang, Libiao Peng, Jiaming Yang, Xiyao Xiao, Sahand Sabour, Xiaohan Zhang, Wenjing Hou, Yijia Zhang, Yuxiao Dong, Jie Tang, and Minlie Huang. 2023{\natexlab{b}}.
\newblock \href {https://arxiv.org/abs/2311.16832} {{{CharacterGLM}}: {{Customizing Chinese Conversational AI Characters}} with {{Large Language Models}}}.
\newblock \emph{ArXiv preprint}, abs/2311.16832.

\end{thebibliography}
\bibliographystyle{acl_natbib}

\clearpage

\appendix

\section{Appendix}
\label{sec:appendix}

\subsection{Examples of Negation and Non-occurrence Scenario Questions}
\label{sec:example_negation_non_occurrence_questions}
\paragraph{Negation Question Type:} For example, ``\chin{在《火影忍者》中，以下哪个不是漩涡鸣人的忍术？\ A. 万象天引 \ B. 影分身之术 \  C. 色诱术 \  D. 螺旋丸}'' (``In \textit{Naruto}, Which of the following is not a ninjutsu of Naruto Uzumaki? A. Universal Pull B. Shadow Clone Jutsu C. Sexy Jutsu D. Rasengan''). The correct answer is A because it is the only ninjutsu that Naruto Uzumaki cannot use among the four given options.

\paragraph{Non-occurrence Scenario Question Type:} For example, ``\chin{在《火影忍者》中，漩涡鸣人是什么时候成为中忍的？\ A. 第四次忍界大战时 \ B. 没有成为中忍 \ C. 佩恩入侵时 \  D. 第七班完成任务后}'' (``In \textit{Naruto}, When did Naruto Uzumaki become a Chūnin? A. During the Fourth Great Ninja War B. Never became a Chūnin C. During Pain's invasion D. After Team 7's mission completion.''). Actually, Naruto never became a Chūnin since he never passed his Chūnin selection exams, even when he became the Seventh Hokage, making the correct choice B.

\subsection{Examples of Different Types of Reasoning Questions}
\label{sec:example_reasoning_questions}
\paragraph{Character Relationship Reasoning:} For example, ``\chin{海莉·比伯的丈夫在《黑衣人3》中客串的什么角色？\ A. 外星人 \ B. 警官 \ C. 医生 \ D. 教师}'' (``What role did Hailey Bieber's husband make a cameo in \textit{Men in Black 3}? A. Alien B. Police Officer C. Doctor D. Teacher''). Models need to first reason out that Hailey Bieber's husband is Justin Bieber, and then find out that Justin Bieber made a cameo appearance as an alien in \textit{Men in Black 3}, so the correct answer is A.

\paragraph{Event Participant Reasoning:} For example, ``\chin{下面哪个人既征服了波斯，又攻下了埃及？\  A. 亚历山大大帝 \  B. 腓力二世 \  C. 冈比西斯二世 \ D. 阿明塔斯三世}'' (``Which of the following people conquered both Persia and Egypt? A. Alexander the Great B. Philip II C. Cambyses II D. Amyntas III''). Models need to know that although more than one person conquered Persia or Egypt (e.g., Cambyses II in option C used to conquer Egypt in 525 BC), Alexander the Great is the only one who conquered both Persia and Egypt among the four options.

\paragraph{Timeline Reasoning:} For example, ``\chin{在《巴黎圣母院》中，哪件事在卡西莫多受鞭笞之刑前发生？\ A. 埃斯梅拉达被判绞刑 \ B. 卡西莫多看见埃斯梅拉达被绞死 \ C. 埃斯梅拉达给卡西莫多喝水 \ D. 卡西莫多把埃斯梅拉达从绞刑架上救下}'' (``In \textit{The Hunchback of Notre Dame}, which event occurs before Quasimodo is whipped? A. Esmeralda was sentenced to hanging B. Quasimodo saw Esmeralda being hanged C. Esmeralda gave Quasimodo water D. Quasimodo took Esmeralda saved from the gallows''). Among these options, only option C happened before whipping, while A, B, and D all happened after that.

\subsection{Zero-shot Experimental Results}
\label{sec:zeroshot_results}

We list zero-shot results by five categories in Table \ref{tab:zh_zeroshot_results} and \ref{tab:en_zeroshot_results}. Overall, the zero-shot performance of most models is slightly lower than (or similar to) five-shot results. We find that although GPT-4 and GPT-3.5 API already have great instruction-following abilities, providing five examples containing characters that are not related to who we aim to assess can still improve both utilization and reasoning of role knowledge. We conjecture that for powerful LLMs like GPT-4 and GPT-3.5, these examples act as cues that activate role knowledge stored in the model. By presenting scenarios where specific roles are depicted, the model can more effectively access and utilize its internal role knowledge, even if these roles are not related to the character that the question actually asks.

\begin{table*}[t]
\large
\centering
\resizebox{\textwidth}{!}{%
\renewcommand*{\arraystretch}{1.1}
\begin{tabular}{@{}c|cccccc|lllllc@{}}
\toprule
\multirow{2}{*}{\textbf{Model}} & \multicolumn{6}{c|}{\textbf{RoleEval-Global (4,000   questions)}}                                                        & \multicolumn{6}{c}{\textbf{RoleEval-Chinese   (2,000 questions)}}                                                                                                                            \\ \cmidrule(l){2-13} 
                                & \textbf{CE}    & \textbf{AC}    & \textbf{MT}    & \textbf{GA}    & \multicolumn{1}{c|}{\textbf{FI}}    & \textbf{Avg.}  & \multicolumn{1}{c}{\textbf{CE}} & \multicolumn{1}{c}{\textbf{AC}} & \multicolumn{1}{c}{\textbf{MT}} & \multicolumn{1}{c}{\textbf{GA}} & \multicolumn{1}{c|}{\textbf{FI}}    & \textbf{Avg.}  \\ \midrule
GPT-3.5-0613                    & 45.25          & 45.88          & 46.38          & 47.12          & \multicolumn{1}{c|}{45.50}          & 46.03          & 41.00                           & 41.00                           & 34.75                           & 33.50                           & \multicolumn{1}{l|}{38.50}          & 37.75          \\
GPT-3.5-1106                    & 45.62          & 45.50          & 45.50          & 44.50          & \multicolumn{1}{c|}{43.25}          & 44.87          & 40.75                           & 41.25                           & 33.25                           & 36.50                           & \multicolumn{1}{l|}{39.50}          & 38.25          \\
GPT-4-0613                      & 70.75          & 71.50          & 73.12          & 69.38          & \multicolumn{1}{c|}{70.50}          & 71.05          & 62.00                           & 59.75                           & 57.25                           & 55.00                           & \multicolumn{1}{l|}{58.75}          & 58.55          \\
GPT-4-1106                      & \textbf{75.12} & \textbf{73.75} & \textbf{75.25} & \textbf{71.50} & \multicolumn{1}{c|}{70.75}          & \textbf{73.27} & \textbf{63.00}                  & 61.25                           & 56.25                           & 62.00                           & \multicolumn{1}{l|}{60.00}          & 60.50          \\
Falcon-7B                       & 23.88          & 25.25          & 25.87          & 28.00          & \multicolumn{1}{c|}{25.12}          & 25.62          & 26.00                           & 24.50                           & 25.50                           & 25.00                           & \multicolumn{1}{l|}{26.25}          & 25.45          \\
Falcon-40B                      & 40.50          & 35.25          & 32.12          & 35.50          & \multicolumn{1}{c|}{45.75}          & 37.82          & 32.25                           & 30.00                           & 36.25                           & 29.75                           & \multicolumn{1}{l|}{31.00}          & 31.85          \\
LLaMA-7B                        & 27.12          & 29.50          & 28.38          & 30.88          & \multicolumn{1}{c|}{25.12}          & 28.20          & 31.50                           & 26.25                           & 26.50                           & 30.50                           & \multicolumn{1}{l|}{28.50}          & 28.65          \\
LLaMA-13B                       & 28.38          & 28.38          & 23.25          & 26.62          & \multicolumn{1}{c|}{30.00}          & 27.33          & 22.50                           & 23.50                           & 25.75                           & 32.00                           & \multicolumn{1}{l|}{26.50}          & 26.05          \\
LLaMA-30B                       & 27.12          & 28.25          & 28.25          & 27.38          & \multicolumn{1}{c|}{23.62}          & 26.92          & 25.00                           & 30.50                           & 23.00                           & 22.75                           & \multicolumn{1}{l|}{32.00}          & 26.65          \\
LLaMA-65B                       & 31.87          & 29.38          & 30.50          & 29.62          & \multicolumn{1}{c|}{30.63}          & 30.40          & 27.25                           & 29.00                           & 28.25                           & 29.00                           & \multicolumn{1}{l|}{29.00}          & 28.50          \\
LLaMA-2-7B                      & 28.38          & 30.75          & 29.25          & 29.88          & \multicolumn{1}{c|}{28.50}          & 29.35          & 27.50                           & 26.00                           & 31.25                           & 27.50                           & \multicolumn{1}{l|}{26.75}          & 27.80          \\
LLaMA-2-13B                     & 30.38          & 30.75          & 32.38          & 29.38          & \multicolumn{1}{c|}{32.50}          & 31.08          & 27.50                           & 27.75                           & 26.50                           & 26.50                           & \multicolumn{1}{l|}{29.50}          & 27.55          \\
LLaMA-2-70B                     & 46.12          & 40.25          & 37.62          & 40.12          & \multicolumn{1}{c|}{42.50}          & 41.32          & 36.50                           & 37.00                           & 34.50                           & 31.00                           & \multicolumn{1}{l|}{34.00}          & 34.60          \\
Mistral-7B                      & 37.00          & 31.37          & 29.00          & 32.12          & \multicolumn{1}{c|}{30.63}          & 32.02          & 32.25                           & 27.50                           & 29.00                           & 26.50                           & \multicolumn{1}{l|}{32.00}          & 29.45          \\ \midrule
MiniMax                         & 53.12          & 56.62          & 60.00          & 56.12          & \multicolumn{1}{c|}{55.00}          & 56.17          & 56.00                           & 52.25                           & 53.00                           & 53.50                           & \multicolumn{1}{l|}{54.25}          & 53.80          \\
Baichuan2-7B                    & 54.37          & 47.25          & 42.50          & 38.38          & \multicolumn{1}{c|}{56.12}          & 47.72          & 40.25                           & 53.25                           & 53.50                           & 49.25                           & \multicolumn{1}{l|}{41.25}          & 47.50          \\
Baichuan2-13B                   & 60.50          & 51.50          & 46.88          & 45.25          & \multicolumn{1}{c|}{57.12}          & 52.25          & 45.75                           & 51.25                           & 55.25                           & 54.50                           & \multicolumn{1}{l|}{44.25}          & 50.20          \\
ChatGLM3-6B                     & 55.38          & 47.62          & 49.12          & 42.00          & \multicolumn{1}{c|}{56.88}          & 50.20          & 45.00                           & 51.25                           & 55.50                           & 49.50                           & \multicolumn{1}{l|}{44.00}          & 49.05          \\
Chinese-LLaMA-2-7B              & 29.12          & 27.25          & 23.25          & 26.75          & \multicolumn{1}{c|}{30.00}          & 27.27          & 22.50                           & 27.00                           & 27.25                           & 31.25                           & \multicolumn{1}{l|}{24.00}          & 26.40          \\
Chinese-LLaMA-2-13B             & 35.62          & 34.50          & 33.62          & 29.25          & \multicolumn{1}{c|}{36.88}          & 33.97          & 31.00                           & 31.75                           & 34.50                           & 31.00                           & \multicolumn{1}{l|}{33.00}          & 32.25          \\
Qwen-7B                         & 56.38          & 43.62          & 46.50          & 38.88          & \multicolumn{1}{c|}{53.87}          & 47.85          & 40.25                           & 49.50                           & 50.75                           & 46.25                           & \multicolumn{1}{l|}{43.25}          & 46.00          \\
Qwen-14B                        & 59.50          & 54.12          & 54.25          & 43.88          & \multicolumn{1}{c|}{58.63}          & 54.08          & 43.25                           & 51.00                           & 58.75                           & 50.75                           & \multicolumn{1}{l|}{50.75}          & 50.90          \\
Qwen-72B                        & 70.62          & 63.75          & 68.38          & 59.25          & \multicolumn{1}{c|}{\textbf{72.00}} & 66.80          & 58.75                           & 66.50                           & \textbf{72.25}                  & \textbf{64.00}                  & \multicolumn{1}{l|}{\textbf{62.75}} & \textbf{64.85} \\
Skywork-13B                     & 58.13          & 50.00          & 47.88          & 43.75          & \multicolumn{1}{c|}{56.25}          & 51.20          & 44.50                           & 53.00                           & 57.50                           & 52.50                           & \multicolumn{1}{l|}{46.00}          & 50.70          \\
Yi-6B                           & 62.75          & 54.25          & 53.00          & 45.12          & \multicolumn{1}{c|}{58.13}          & 54.65          & 46.00                           & 57.50                           & 62.00                           & 58.50                           & \multicolumn{1}{l|}{45.25}          & 53.85          \\
Yi-34B                          & 67.38          & 59.00          & 62.75          & 51.62          & \multicolumn{1}{c|}{67.88}          & 61.73          & 54.75                           & \textbf{67.25}                  & 68.75                           & 59.75                           & \multicolumn{1}{l|}{56.00}          & 61.30          \\ \bottomrule
\end{tabular}%
}
\caption{Zero-shot results on RoleEval (\textbf{zh}) in five categories: celebrities (\textbf{CE}), anime and comics (\textbf{AC}), movie and TV series (\textbf{MT}), games (\textbf{GA}), and fiction (\textbf{FI}).}
\label{tab:zh_zeroshot_results}
\end{table*}

\begin{table*}[t]
\large
\centering
\resizebox{\textwidth}{!}{%
\renewcommand*{\arraystretch}{1.1}
\begin{tabular}{@{}c|lllllc|lllllc@{}}
\toprule
\multirow{2}{*}{\textbf{Model}} & \multicolumn{6}{c|}{\textbf{RoleEval-Global (4,000   questions)}}                                                                                                                            & \multicolumn{6}{c}{\textbf{RoleEval-Chinese   (2,000 questions)}}                                                                                                                            \\ \cmidrule(l){2-13} 
                                & \multicolumn{1}{c}{\textbf{CE}} & \multicolumn{1}{c}{\textbf{AC}} & \multicolumn{1}{c}{\textbf{MT}} & \multicolumn{1}{c}{\textbf{GA}} & \multicolumn{1}{c|}{\textbf{FI}}    & \textbf{Avg.}  & \multicolumn{1}{c}{\textbf{CE}} & \multicolumn{1}{c}{\textbf{AC}} & \multicolumn{1}{c}{\textbf{MT}} & \multicolumn{1}{c}{\textbf{GA}} & \multicolumn{1}{c|}{\textbf{FI}}    & \textbf{Avg.}  \\ \midrule
GPT-3.5-0613                    & 57.00                           & 59.00                           & 56.12                           & 59.13                           & \multicolumn{1}{l|}{57.50}          & 57.75          & 37.75                           & 44.50                           & 43.75                           & 43.25                           & \multicolumn{1}{l|}{45.00}          & 42.85          \\
GPT-3.5-1106                    & 55.50                           & 53.62                           & 53.50                           & 56.00                           & \multicolumn{1}{l|}{53.50}          & 54.42          & 39.25                           & 40.50                           & 42.00                           & 39.00                           & \multicolumn{1}{l|}{42.00}          & 40.55          \\
GPT-4-0613                      & 74.38                           & 74.12                           & \textbf{73.88}                  & 72.75                           & \multicolumn{1}{l|}{71.25}          & 73.28          & 48.50                           & 56.00                           & \textbf{60.50}                  & 58.75                           & \multicolumn{1}{l|}{55.25}          & 55.80          \\
GPT-4-1106                      & \textbf{74.38}                  & \textbf{78.62}                  & 72.38                           & \textbf{74.38}                  & \multicolumn{1}{l|}{\textbf{73.00}} & \textbf{74.55} & 51.00                           & \textbf{56.75}                  & 59.75                           & \textbf{60.25}                  & \multicolumn{1}{l|}{\textbf{57.75}} & \textbf{57.10} \\
Falcon-7B                       & 28.38                           & 26.75                           & 23.12                           & 26.12                           & \multicolumn{1}{l|}{31.37}          & 27.15          & 26.50                           & 24.50                           & 30.75                           & 28.75                           & \multicolumn{1}{l|}{24.00}          & 26.90          \\
Falcon-40B                      & 47.12                           & 43.88                           & 47.12                           & 42.38                           & \multicolumn{1}{l|}{48.88}          & 45.88          & 37.50                           & 36.75                           & 30.25                           & 38.25                           & \multicolumn{1}{l|}{36.75}          & 35.90          \\
LLaMA-7B                        & 32.00                           & 31.50                           & 29.38                           & 31.00                           & \multicolumn{1}{l|}{32.88}          & 31.35          & 23.00                           & 24.75                           & 29.25                           & 30.25                           & \multicolumn{1}{l|}{29.25}          & 27.30          \\
LLaMA-13B                       & 35.88                           & 36.50                           & 37.00                           & 32.25                           & \multicolumn{1}{l|}{42.25}          & 36.78          & 33.50                           & 28.25                           & 29.25                           & 34.75                           & \multicolumn{1}{l|}{30.75}          & 31.30          \\
LLaMA-30B                       & 45.62                           & 41.62                           & 47.12                           & 41.88                           & \multicolumn{1}{l|}{51.00}          & 45.45          & 35.75                           & 31.75                           & 35.00                           & 38.00                           & \multicolumn{1}{l|}{36.00}          & 35.30          \\
LLaMA-65B                       & 53.62                           & 49.75                           & 50.88                           & 41.75                           & \multicolumn{1}{l|}{54.75}          & 50.15          & 38.00                           & 35.25                           & 35.00                           & 40.00                           & \multicolumn{1}{l|}{37.25}          & 37.10          \\
LLaMA-2-7B                      & 33.62                           & 32.88                           & 31.87                           & 33.88                           & \multicolumn{1}{l|}{37.50}          & 33.95          & 25.25                           & 27.00                           & 30.50                           & 29.25                           & \multicolumn{1}{l|}{31.75}          & 28.75          \\
LLaMA-2-13B                     & 48.88                           & 45.62                           & 48.75                           & 42.50                           & \multicolumn{1}{l|}{47.50}          & 46.65          & 33.25                           & 38.25                           & 35.50                           & 39.75                           & \multicolumn{1}{l|}{29.75}          & 35.30          \\
LLaMA-2-70B                     & 58.50                           & 54.37                           & 57.75                           & 47.88                           & \multicolumn{1}{l|}{61.00}          & 55.90          & 37.75                           & 41.50                           & 39.25                           & 43.50                           & \multicolumn{1}{l|}{42.50}          & 40.90          \\
Mistral-7B                      & 53.87                           & 48.38                           & 48.50                           & 44.75                           & \multicolumn{1}{l|}{50.12}          & 49.12          & 37.50                           & 36.00                           & 33.25                           & 38.50                           & \multicolumn{1}{l|}{36.00}          & 36.25          \\ \midrule
MiniMax                         & 54.50                           & 55.88                           & 52.50                           & 54.12                           & \multicolumn{1}{l|}{53.00}          & 54.00          & 36.75                           & 40.50                           & 41.00                           & 40.25                           & \multicolumn{1}{l|}{40.50}          & 39.80          \\
Baichuan2-7B                    & 52.12                           & 43.75                           & 44.88                           & 39.62                           & \multicolumn{1}{l|}{49.00}          & 45.87          & 35.25                           & 34.00                           & 30.00                           & 38.50                           & \multicolumn{1}{l|}{34.25}          & 34.40          \\
Baichuan2-13B                   & 55.00                           & 47.38                           & 48.12                           & 42.62                           & \multicolumn{1}{l|}{54.25}          & 49.47          & 39.00                           & 38.50                           & 30.75                           & 40.75                           & \multicolumn{1}{l|}{35.50}          & 36.90          \\
ChatGLM3-6B                     & 52.12                           & 46.75                           & 51.62                           & 43.75                           & \multicolumn{1}{l|}{52.62}          & 49.37          & 34.75                           & 37.50                           & 35.00                           & 41.00                           & \multicolumn{1}{l|}{39.75}          & 37.60          \\
Chinese-LLaMA-2-7B              & 29.00                           & 26.62                           & 22.38                           & 25.50                           & \multicolumn{1}{l|}{30.25}          & 26.75          & 23.75                           & 22.75                           & 31.75                           & 26.75                           & \multicolumn{1}{l|}{26.75}          & 26.35          \\
Chinese-LLaMA-2-13B             & 41.00                           & 43.38                           & 43.75                           & 40.25                           & \multicolumn{1}{l|}{45.38}          & 42.75          & 31.50                           & 33.00                           & 33.00                           & 35.00                           & \multicolumn{1}{l|}{33.25}          & 33.15          \\
Qwen-7B                         & 53.87                           & 43.88                           & 49.00                           & 40.50                           & \multicolumn{1}{l|}{50.88}          & 47.63          & 35.25                           & 31.75                           & 36.25                           & 39.75                           & \multicolumn{1}{l|}{35.75}          & 35.75          \\
Qwen-14B                        & 59.62                           & 50.88                           & 54.37                           & 44.62                           & \multicolumn{1}{l|}{55.88}          & 53.07          & 42.25                           & 42.25                           & 37.50                           & 44.50                           & \multicolumn{1}{l|}{41.75}          & 41.65          \\
Qwen-72B                        & 69.00                           & 62.25                           & 67.88                           & 57.12                           & \multicolumn{1}{l|}{68.12}          & 64.87          & 48.00                           & 47.00                           & 46.25                           & 51.25                           & \multicolumn{1}{l|}{55.00}          & 49.50          \\
Skywork-13B                     & 53.62                           & 46.25                           & 50.25                           & 41.38                           & \multicolumn{1}{l|}{53.62}          & 49.02          & 36.75                           & 36.50                           & 33.25                           & 42.00                           & \multicolumn{1}{l|}{37.50}          & 37.20          \\
Yi-6B                           & 57.38                           & 51.50                           & 52.88                           & 43.62                           & \multicolumn{1}{l|}{53.25}          & 51.73          & 39.75                           & 39.50                           & 37.25                           & 42.00                           & \multicolumn{1}{l|}{42.00}          & 40.10          \\
Yi-34B                          & 67.00                           & 61.62                           & 63.00                           & 54.00                           & \multicolumn{1}{l|}{62.50}          & 61.62          & \textbf{52.25}                  & 50.00                           & 39.50                           & 51.75                           & \multicolumn{1}{l|}{49.25}          & 48.55          \\ \bottomrule
\end{tabular}%
}
\caption{Zero-shot results on RoleEval (\textbf{en}) in five categories: celebrities (\textbf{CE}), anime and comics (\textbf{AC}), movie and TV series (\textbf{MT}), games (\textbf{GA}), and fiction (\textbf{FI}).}
\label{tab:en_zeroshot_results}
\end{table*}

\subsection{Results by Knowledge and Reasoning}
\label{sec:results_by_knowledge_and_reasoning}
We provide results of knowledge and reasoning questions for RoleEval in Table \ref{tab:RoleEval_zh_knowledge_reasoning} and \ref{tab:RoleEval_en_knowledge_reasoning}. We observe a clear positive correlation between the breadth of knowledge and the correctness of reasoning. Among these LLMs, the GPT-4 model shows both extensive knowledge and robust reasoning capability. However, it is noteworthy that in specific versions of GPT-4 API, the \texttt{gpt-4-1106} model demonstrates a more comprehensive knowledge base, while the \texttt{gpt-4-0613} model is better at reasoning. This delineation suggests that while there is a general trend of knowledge and reasoning abilities advancing in tandem, individual models may specialize or excel in one aspect over the other.

Further, the experimental findings in LLMs trained for different languages reveal intriguing insights. In English LLMs, Mistral-7B has shown remarkable reasoning capabilities, closely rivaling those of the LLaMA-65B model. This performance indicates a significant advancement in reasoning, as smaller models approach the upper echelons of reasoning skills previously dominated by more advanced models. In the realm of Chinese LLMs, Qwen-72B and Yi-34B also stand out with their exceptional reasoning abilities, positioning themselves between the GPT-3.5 and GPT-4 models. This not only underscores the progress in developing LLMs in different languages but also indicates the potential for LLMs to improve both knowledge and reasoning by considering a wider range of languages when constructing training corpora.

\begin{figure*}[h]
    \centering
    \begin{subfigure}[b]{0.45\textwidth}
        \includegraphics[width=\textwidth]{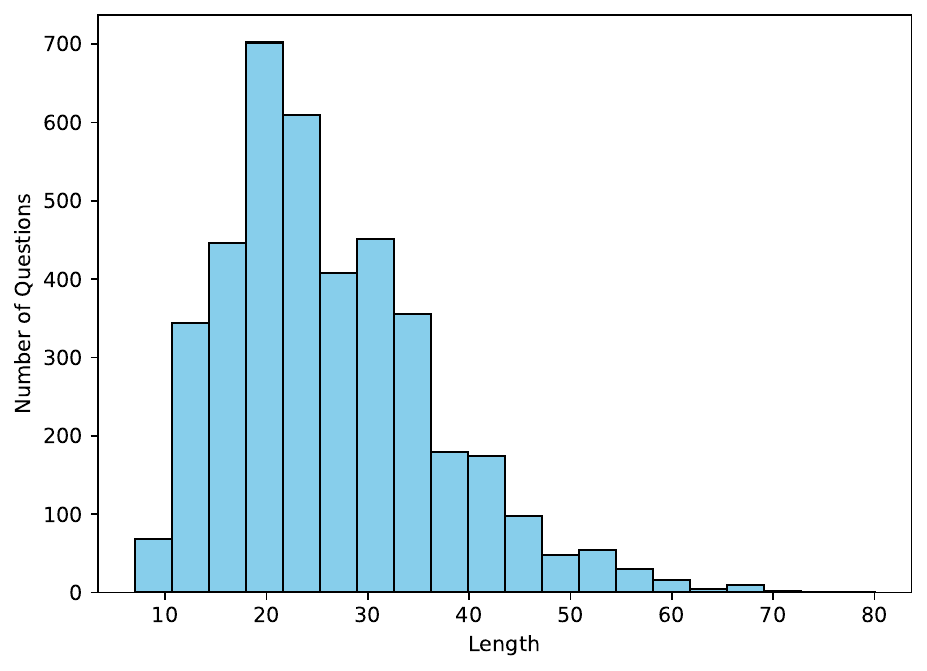}
        \caption{\textit{RoleEval-Global} (\textbf{zh})}
        \label{fig:global_zh_len}
    \end{subfigure}
    \hfill
    \begin{subfigure}[b]{0.45\textwidth}
        \includegraphics[width=\textwidth]{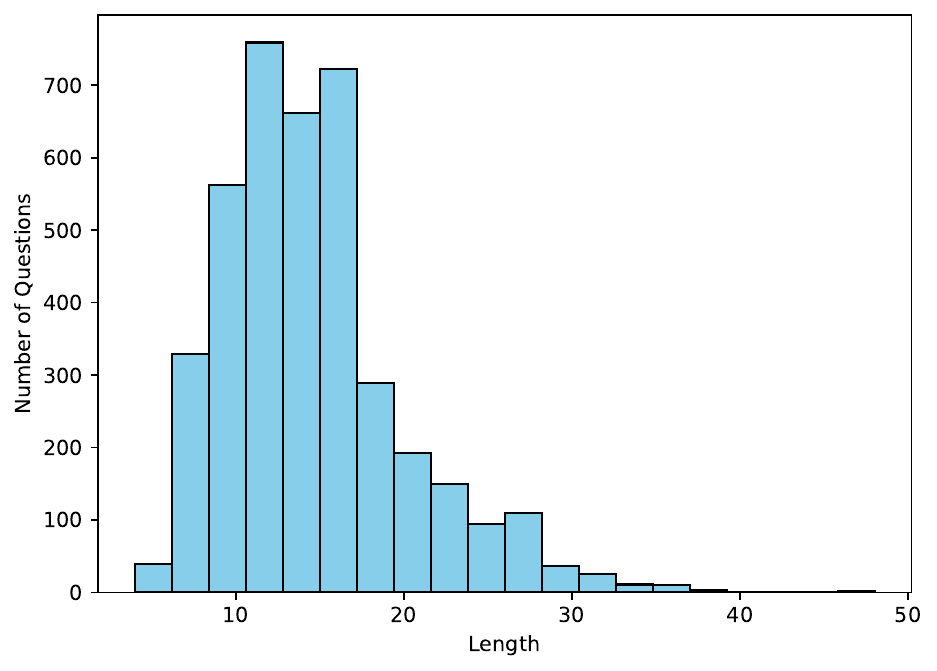}
        \caption{\textit{RoleEval-Global} (\textbf{en})}
        \label{fig:global_en_len}
    \end{subfigure}
    \newline
    \begin{subfigure}[b]{0.45\textwidth}
        \includegraphics[width=\textwidth]{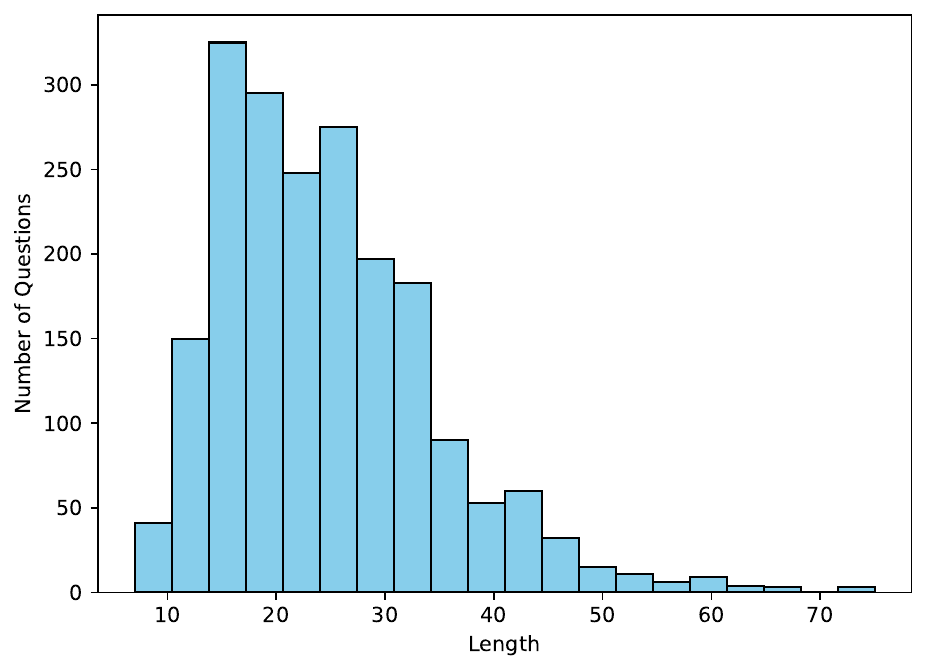}
        \caption{\textit{RoleEval-Chinese} (\textbf{zh})}
        \label{fig:chinese_zh_len}
    \end{subfigure}
    \hfill
    \begin{subfigure}[b]{0.45\textwidth}
        \includegraphics[width=\textwidth]{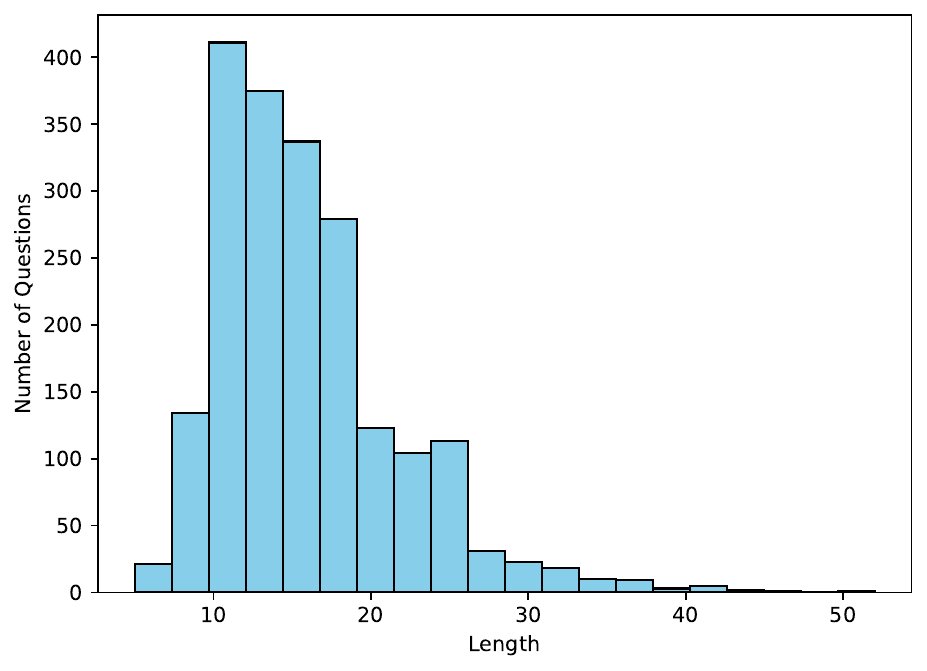}
        \caption{\textit{RoleEval-Chinese} (\textbf{en})}
        \label{fig:chinese_en_len}
    \end{subfigure}
    \caption{Visualization of the length distribution of questions in RoleEval.}
    \label{fig:RoleEval_length}
\end{figure*}

\begin{table*}[]
\large
\centering
\resizebox{\textwidth}{!}{%
\renewcommand*{\arraystretch}{1.1}
\begin{tabular}{@{}c|cccc|cccc@{}}
\toprule
\multirow{3}{*}{\textbf{Model}} & \multicolumn{4}{c|}{\textbf{RoleEval-Global (4,000 questions)}}                                                       & \multicolumn{4}{c}{\textbf{RoleEval-Chinese (2,000 questions)}}                                                      \\ \cmidrule(l){2-9} 
                                & \multicolumn{2}{c|}{\textbf{Knowledge (3,400 questions)}}   & \multicolumn{2}{c|}{\textbf{Reasoning (600 questions)}} & \multicolumn{2}{c|}{\textbf{Knowledge (1,700 questions)}}   & \multicolumn{2}{c}{\textbf{Reasoning (300 questions)}} \\ \cmidrule(l){2-9} 
                                & \textbf{Zero-shot} & \multicolumn{1}{c|}{\textbf{Few-shot}} & \textbf{Zero-shot}          & \textbf{Few-shot}         & \textbf{Zero-shot} & \multicolumn{1}{c|}{\textbf{Few-shot}} & \textbf{Zero-shot}         & \textbf{Few-shot}         \\ \midrule
GPT-3.5-0613                    & 46.06              & \multicolumn{1}{c|}{48.47}             & 45.83                       & 50.17                     & 39.06              & \multicolumn{1}{c|}{42.47}             & 30.33                      & 37.00                     \\
GPT-3.5-1106                    & 44.03              & \multicolumn{1}{c|}{49.53}             & 49.67                       & 52.83                     & 38.65              & \multicolumn{1}{c|}{45.12}             & 36.00                      & 37.00                     \\
GPT-4-0613                      & 71.09              & \multicolumn{1}{c|}{72.74}             & \textbf{70.83}              & \textbf{70.00}            & 58.94              & \multicolumn{1}{c|}{59.65}             & 56.33                      & 53.67                     \\
GPT-4-1106                      & \textbf{73.76}     & \multicolumn{1}{c|}{\textbf{73.97}}    & 70.50                       & \textbf{70.00}                   & 61.00              & \multicolumn{1}{c|}{63.59}             & 57.67                      & 58.00                     \\
Falcon-7B                       & 25.91              & \multicolumn{1}{c|}{27.06}             & 24.00                       & 23.50                     & 25.88              & \multicolumn{1}{c|}{28.53}             & 23.00                      & 27.33                     \\
Falcon-40B                      & 37.32              & \multicolumn{1}{c|}{35.65}             & 40.67                       & 37.00                     & 32.47              & \multicolumn{1}{c|}{32.29}             & 28.33                      & 29.33                     \\
LLaMA-7B                        & 28.74              & \multicolumn{1}{c|}{28.12}             & 25.17                       & 24.83                     & 29.12              & \multicolumn{1}{c|}{26.53}             & 26.00                      & 23.67                     \\
LLaMA-13B                       & 26.29              & \multicolumn{1}{c|}{28.03}             & 33.17                       & 27.17                     & 25.76              & \multicolumn{1}{c|}{28.41}             & 27.67                      & 24.67                     \\
LLaMA-30B                       & 27.38              & \multicolumn{1}{c|}{28.24}             & 24.33                       & 30.17                     & 26.53              & \multicolumn{1}{c|}{28.82}             & 27.33                      & 29.33                     \\
LLaMA-65B                       & 29.94              & \multicolumn{1}{c|}{32.62}             & 33.00                       & 32.00                     & 27.82              & \multicolumn{1}{c|}{31.76}             & 32.33                      & 27.67                     \\
LLaMA-2-7B                      & 29.59              & \multicolumn{1}{c|}{33.38}             & 28.00                       & 35.33                     & 28.06              & \multicolumn{1}{c|}{30.47}             & 26.33                      & 29.67                     \\
LLaMA-2-13B                     & 31.29              & \multicolumn{1}{c|}{33.41}             & 29.83                       & 33.50                     & 27.59              & \multicolumn{1}{c|}{28.59}             & 27.33                      & 28.00                     \\
LLaMA-2-70B                     & 41.18              & \multicolumn{1}{c|}{44.47}             & 42.17                       & 46.00                     & 35.18              & \multicolumn{1}{c|}{36.65}             & 31.33                      & 34.00                     \\
Mistral-7B                      & 32.32              & \multicolumn{1}{c|}{33.15}             & 30.33                       & 34.67                     & 30.06              & \multicolumn{1}{c|}{32.35}             & 26.00                      & 31.33                     \\ \midrule
MiniMax                         & 54.82              & \multicolumn{1}{c|}{54.29}             & 63.83                       & 63.50                     & 52.94              & \multicolumn{1}{c|}{54.00}             & 58.67                      & 58.33                     \\
Baichuan2-7B                    & 46.76              & \multicolumn{1}{c|}{48.32}             & 53.17                       & 51.50                     & 48.18              & \multicolumn{1}{c|}{49.59}             & 43.67                      & 48.67                     \\
Baichuan2-13B                   & 52.35              & \multicolumn{1}{c|}{54.03}             & 51.67                       & 55.50                     & 51.24              & \multicolumn{1}{c|}{53.41}             & 44.33                      & 49.33                     \\
ChatGLM3-6B                     & 49.82              & \multicolumn{1}{c|}{48.79}             & 52.33                       & 55.33                     & 49.18              & \multicolumn{1}{c|}{48.65}             & 48.33                      & 50.67                     \\
Chinese-LLaMA-2-7B              & 26.62              & \multicolumn{1}{c|}{34.94}             & 31.00                       & 39.00                     & 26.12              & \multicolumn{1}{c|}{33.53}             & 28.00                      & 27.33                     \\
Chinese-LLaMA-2-13B             & 33.12              & \multicolumn{1}{c|}{39.06}             & 38.83                       & 42.00                     & 31.88              & \multicolumn{1}{c|}{35.71}             & 34.33                      & 39.67                     \\
Qwen-7B                         & 47.47              & \multicolumn{1}{c|}{47.76}             & 50.00                       & 48.67                     & 45.88              & \multicolumn{1}{c|}{46.41}             & 46.67                      & 49.67                     \\
Qwen-14B                        & 53.79              & \multicolumn{1}{c|}{54.68}             & 55.67                       & 54.67                     & 51.47              & \multicolumn{1}{c|}{53.18}             & 47.67                      & 51.67                     \\
Qwen-72B                        & 67.18              & \multicolumn{1}{c|}{67.59}             & 64.67                       & 66.83                     & \textbf{65.76}     & \multicolumn{1}{c|}{\textbf{67.00}}    & \textbf{59.67}             & 61.67                     \\
Skywork-13B                     & 50.03              & \multicolumn{1}{c|}{52.44}             & 57.83                       & 57.50                     & 51.00              & \multicolumn{1}{c|}{52.76}             & 49.00                      & 51.67                     \\
Yi-6B                           & 55.18              & \multicolumn{1}{c|}{54.09}             & 51.67                       & 55.83                     & 55.53              & \multicolumn{1}{c|}{55.88}             & 44.33                      & 52.00                     \\
Yi-34B                          & 62.79              & \multicolumn{1}{c|}{65.38}             & 55.67                       & 68.33                     & 62.29              & \multicolumn{1}{c|}{64.71}             & 55.67                      & \textbf{64.00}            \\ \bottomrule
\end{tabular}%
}
\caption{Results by knowledge and reasoning questions in RoleEval (\textbf{zh}).}
\label{tab:RoleEval_zh_knowledge_reasoning}
\end{table*}

\begin{table*}[]
\large
\centering
\resizebox{\textwidth}{!}{%
\renewcommand*{\arraystretch}{1.1}
\begin{tabular}{@{}c|cccc|cccc@{}}
\toprule
\multirow{3}{*}{\textbf{Model}} & \multicolumn{4}{c|}{\textbf{RoleEval-Global (4,000 questions)}}                                                       & \multicolumn{4}{c}{\textbf{RoleEval-Chinese (2,000 questions)}}                                                      \\ \cmidrule(l){2-9} 
                                & \multicolumn{2}{c|}{\textbf{Knowledge (3,400 questions)}}   & \multicolumn{2}{c|}{\textbf{Reasoning (600 questions)}} & \multicolumn{2}{c|}{\textbf{Knowledge (1,700 questions)}}   & \multicolumn{2}{c}{\textbf{Reasoning (300 questions)}} \\ \cmidrule(l){2-9} 
                                & \textbf{Zero-shot} & \multicolumn{1}{c|}{\textbf{Few-shot}} & \textbf{Zero-shot}          & \textbf{Few-shot}         & \textbf{Zero-shot} & \multicolumn{1}{c|}{\textbf{Few-shot}} & \textbf{Zero-shot}         & \textbf{Few-shot}         \\ \midrule
GPT-3.5-0613                    & 57.03              & \multicolumn{1}{c|}{57.88}             & 61.83                       & 61.50                     & 42.71              & \multicolumn{1}{c|}{43.47}             & 43.67                      & 47.00                     \\
GPT-3.5-1106                    & 53.15              & \multicolumn{1}{c|}{55.82}             & 61.67                       & 62.50                     & 39.65              & \multicolumn{1}{c|}{43.18}             & 45.67                      & 46.33                     \\
GPT-4-0613                      & 72.94              & \multicolumn{1}{c|}{75.79}             & \textbf{75.17}              & \textbf{77.33}            & 55.65              & \multicolumn{1}{c|}{60.65}             & \textbf{56.67}             & \textbf{63.00}            \\
GPT-4-1106                      & \textbf{74.62}     & \multicolumn{1}{c|}{\textbf{76.29}}    & 74.17                       & 74.33                     & \textbf{57.82}     & \multicolumn{1}{c|}{\textbf{60.71}}    & 53.00                      & 58.00                     \\
Falcon-7B                       & 26.50              & \multicolumn{1}{c|}{28.68}             & 30.83                       & 28.00                     & 27.00              & \multicolumn{1}{c|}{28.18}             & 26.33                      & 28.33                     \\
Falcon-40B                      & 45.53              & \multicolumn{1}{c|}{46.71}             & 47.83                       & 48.83                     & 36.65              & \multicolumn{1}{c|}{36.41}             & 31.67                      & 29.67                     \\
LLaMA-7B                        & 30.74              & \multicolumn{1}{c|}{29.41}             & 34.83                       & 34.83                     & 27.06              & \multicolumn{1}{c|}{28.06}             & 28.67                      & 32.33                     \\
LLaMA-13B                       & 35.47              & \multicolumn{1}{c|}{39.35}             & 44.17                       & 47.83                     & 30.88              & \multicolumn{1}{c|}{32.47}             & 33.67                      & 37.00                     \\
LLaMA-30B                       & 44.47              & \multicolumn{1}{c|}{47.97}             & 51.00                       & 52.00                     & 34.88              & \multicolumn{1}{c|}{35.18}             & 37.67                      & 35.67                     \\
LLaMA-65B                       & 49.41              & \multicolumn{1}{c|}{52.62}             & 54.33                       & 55.33                     & 37.00              & \multicolumn{1}{c|}{38.94}             & 37.67                      & 38.33                     \\
LLaMA-2-7B                      & 33.32              & \multicolumn{1}{c|}{38.59}             & 37.50                       & 44.50                     & 28.76              & \multicolumn{1}{c|}{31.82}             & 28.67                      & 33.67                     \\
LLaMA-2-13B                     & 46.44              & \multicolumn{1}{c|}{46.35}             & 47.83                       & 46.50                     & 35.82              & \multicolumn{1}{c|}{34.00}             & 32.33                      & 34.00                     \\
LLaMA-2-70B                     & 55.41              & \multicolumn{1}{c|}{58.41}             & 58.67                       & 59.50                     & 41.76              & \multicolumn{1}{c|}{43.65}             & 36.00                      & 40.67                     \\
Mistral-7B                      & 48.26              & \multicolumn{1}{c|}{48.50}             & 54.00                       & 55.50                     & 36.65              & \multicolumn{1}{c|}{36.12}             & 34.00                      & 36.67                     \\ \midrule
MiniMax                         & 52.44              & \multicolumn{1}{c|}{52.79}             & 62.83                       & 61.17                     & 38.76              & \multicolumn{1}{c|}{38.06}             & 45.67                      & 39.67                     \\
Baichuan2-7B                    & 44.79              & \multicolumn{1}{c|}{46.71}             & 52.00                       & 51.67                     & 33.71              & \multicolumn{1}{c|}{35.88}             & 38.33                      & 41.00                     \\
Baichuan2-13B                   & 48.97              & \multicolumn{1}{c|}{50.21}             & 52.33                       & 55.17                     & 36.94              & \multicolumn{1}{c|}{36.53}             & 36.67                      & 33.33                     \\
ChatGLM3-6B                     & 48.38              & \multicolumn{1}{c|}{48.50}             & 55.00                       & 54.33                     & 37.35              & \multicolumn{1}{c|}{38.59}             & 39.00                      & 39.33                     \\
Chinese-LLaMA-2-7B              & 25.82              & \multicolumn{1}{c|}{34.59}             & 32.00                       & 37.00                     & 26.00              & \multicolumn{1}{c|}{29.41}             & 28.33                      & 32.67                     \\
Chinese-LLaMA-2-13B             & 41.62              & \multicolumn{1}{c|}{46.21}             & 49.17                       & 50.17                     & 33.12              & \multicolumn{1}{c|}{34.12}             & 33.33                      & 35.67                     \\
Qwen-7B                         & 47.32              & \multicolumn{1}{c|}{47.35}             & 49.33                       & 50.67                     & 35.53              & \multicolumn{1}{c|}{38.24}             & 37.00                      & 37.67                     \\
Qwen-14B                        & 52.44              & \multicolumn{1}{c|}{52.97}             & 56.67                       & 53.83                     & 41.35              & \multicolumn{1}{c|}{40.41}             & 43.33                      & 40.67                     \\
Qwen-72B                        & 63.91              & \multicolumn{1}{c|}{64.79}             & 70.33                       & 68.00                     & 49.47              & \multicolumn{1}{c|}{50.82}             & 49.67                      & 47.33                     \\
Skywork-13B                     & 47.26              & \multicolumn{1}{c|}{49.44}             & 59.00                       & 56.67                     & 36.94              & \multicolumn{1}{c|}{38.59}             & 38.67                      & 37.67                     \\
Yi-6B                           & 51.03              & \multicolumn{1}{c|}{53.35}             & 55.67                       & 56.50                     & 40.00              & \multicolumn{1}{c|}{42.82}             & 40.67                      & 39.33                     \\
Yi-34B                          & 61.24              & \multicolumn{1}{c|}{64.94}             & 63.83                       & 68.17                     & 48.53              & \multicolumn{1}{c|}{53.82}             & 48.67                      & 51.67                     \\ \bottomrule
\end{tabular}%
}
\caption{Results by knowledge and reasoning questions in RoleEval (\textbf{en}).}
\label{tab:RoleEval_en_knowledge_reasoning}
\end{table*}

\subsection{Detailed Character List}
\label{sec:character_list}

We list all the collected 300 characters in Table \ref{tab:celebrities_char_list}, \ref{tab:anime_comics_char_list}, \ref{tab:movie_tv_char_list}, \ref{tab:game_char_list}, \ref{tab:fiction_char_list} and \ref{tab:RoleEval_chinese_list}, with both the name and source of each character in Chinese and English. 

\begin{table*}[h]
    \centering
    \resizebox{\textwidth}{!}{%
        \begin{tabular}{c|c|c|c}
            \hline
            \textbf{Name} & \textbf{Name} & \textbf{Name} & \textbf{Name} \\ \hline
            \chin{爱因斯坦} (Albert Einstein) & \chin{安徒生} (Hans Christian Andersen) & \chin{李相赫} (Lee Sang-hyeok, known as Faker) & \chin{苏格拉底} (Socrates) \\ 
            \chin{亚历山大大帝} (Alexander the Great) & \chin{村上春树} (Haruki Murakami) & \chin{达芬奇} (Leonardo da Vinci) & \chin{史蒂夫·乔布斯} (Steve Jobs) \\ 
            \chin{亚里士多德} (Aristotle) & \chin{宫崎骏} (Hayao Miyazaki) & \chin{梅西} (Lionel Messi) & \chin{西尔维斯特·史泰龙} (Sylvester Stallone) \\ 
            \chin{奥黛丽·赫本} (Audrey Hepburn) & \chin{成龙} (Jackie Chan) & \chin{鲁迅} (Lu Xun) & \chin{泰勒·斯威夫特} (Taylor Swift) \\ 
            \chin{比尔·盖茨} (Bill Gates) & \chin{贾斯汀·比伯} (Justin Bieber) & \chin{甘地} (Mahatma Gandhi) & \chin{托马斯·爱迪生} (Thomas Edison) \\ 
            \chin{李小龙} (Bruce Lee) & \chin{马克思} (Karl Marx) & \chin{玛丽莲·梦露} (Marilyn Monroe) & \chin{汤姆·汉克斯} (Tom Hanks) \\ 
            \chin{达尔文} (Charles Darwin) & \chin{金·卡戴珊} (Kim Kardashian) & \chin{迈克尔·杰克逊} (Michael Jackson) & \chin{梵高} (Vincent van Gogh) \\ 
            \chin{马斯克} (Elon Musk) & \chin{孔子} (Kongzi, known as Confucius) & \chin{拿破仑} (Napoleon) & \chin{莎士比亚} (William Shakespeare) \\ 
            \chin{埃尔维斯·普雷斯利} (Elvis Presley) & \chin{权志龙} (Kwon Ji-yong, known as G-Dragon) & \chin{曼德拉} (Nelson Mandela) & \chin{姚明} (Yao Ming) \\ 
            \chin{成吉思汗} (Genghis Khan) & \chin{郎朗} (Lang Lang) & \chin{伊丽莎白二世} (Queen Elizabeth II) & \chin{羽生结弦} (Yuzuru Hanyu) \\ \hline
        \end{tabular}
    }
    \caption{Characters from real-world celebrities in \textit{RoleEval-Global}.}
    \label{tab:celebrities_char_list}
\end{table*}

\begin{table*}[]
    \centering
    \resizebox{\textwidth}{!}{%
        \begin{tabular}{cc|cc}
        \hline
        \textbf{Name}                    & \textbf{Source}                                         & \textbf{Name}                        & \textbf{Source}                                              \\ \hline
        \chin{阿姆罗·雷} (Amuro Ray)         & \chin{机动战士高达} (\textit{Mobile Suit Gundam})             & \chin{蒙奇·D·路飞} (Monkey D. Luffy)     & \chin{海贼王} (\textit{One Piece})                              \\
        \chin{花花} (Blossom)              & \chin{飞天小女警} (\textit{Powerpuff Girls})                 & \chin{墨菲斯} (Morpheus)                & \chin{睡魔} (\textit{The Sandman})                             \\
        \chin{樱桃子} (Chibi Maruko-chan)   & \chin{樱桃小丸子} (\textit{Chibi Maruko-chan})               & \chin{草}\abchin{薙}\chin{素子} (Motoko Kusanagi)        & \chin{攻壳机动队} (\textit{Ghost in the Shell})                   \\
        \chin{江户川柯南} (Conan Edogawa)     & \chin{名侦探柯南} (\textit{Detective Conan})                 & \chin{漩涡鸣人} (Naruto Uzumaki)         & \chin{火影忍者} (\textit{Naruto})                                \\
        \chin{唐老鸭} (Donald Duck)         & \chin{米老鼠和唐老鸭} (\textit{Mickey Mouse and Donald Duck})  & \chin{绫波丽} (Rei Ayanami)             & \chin{新世纪福音战士} (\textit{Neon Genesis Evangelion})            \\
        \chin{哆啦A梦} (Doraemon)           & \chin{哆啦A梦} (\textit{Doraemon})                         & \chin{瑞克} (Rick Sanchez)             & \chin{瑞克和莫蒂} (\textit{Rick and Morty})                       \\
        \chin{艾伦·耶格尔} (Eren Yeager)      & \chin{进击的巨人} (\textit{Attack on Titan})                 & \chin{越前龙马} (Ryoma Echizen)          & \chin{网球王子} (\textit{The Prince of Tennis})                  \\
        \chin{菲·瓦伦坦} (Faye Valentine)    & \chin{星际牛仔} (\textit{Cowboy Bebop})                     & \chin{木之本樱} (Sakura Kinomoto)        & \chin{魔卡少女樱} (\textit{Cardcaptor Sakura})                    \\
        \chin{加菲猫} (Garfield)            & \chin{加菲猫} (\textit{Garfield})                          & \chin{五条悟} (Satoru Gojo)             & \chin{咒术回战} (\textit{Jujutsu Kaisen})                        \\
        \chin{坂田银时} (Gintoki Sakata)     & \chin{银魂} (\textit{Gintama})                            & \chin{野原新之助} (Shinnosuke Nohara)     & \chin{蜡笔小新} (\textit{Crayon Shin-chan})                      \\
        \chin{乔鲁诺·乔巴纳} (Giorno Giovanna) & \chin{JOJO的奇妙冒险} (\textit{JoJo's Bizarre Adventure})    & \chin{史努比} (Snoopy)                  & \chin{花生漫画} (\textit{Peanuts})                               \\
        \chin{樱木花道} (Hanamichi Sakuragi) & \chin{灌篮高手} (\textit{Slam Dunk})                        & \chin{白雪公主} (Snow White)             & \chin{白雪公主和七个小矮人} (\textit{Snow White and the Seven Dwarfs}) \\
        \chin{地狱男爵} (Hellboy)            & \chin{地狱男爵系列} (\textit{Hellboy Series})                 & \chin{海绵宝宝} (SpongeBob SquarePants)  & \chin{海绵宝宝} (\textit{SpongeBob SquarePants})                 \\
        \chin{黑崎一护} (Ichigo Kurosaki)    & \chin{BLEACH} (\textit{Bleach})                         & \chin{超人} (Superman)                 & \chin{超人} (\textit{Superman})                                \\
        \chin{杰瑞} (Jerry)                & \chin{猫和老鼠} (\textit{Tom and Jerry})                    & \chin{夏目贵志} (Takashi Natsume)        & \chin{夏目友人帐} (\textit{Natsume's Book of Friends})            \\
        \chin{桔梗} (Kikyo)                & \chin{犬夜叉} (\textit{Inuyasha})                          & \chin{天津饭} (Tien Shinhan)            & \chin{龙珠} (\textit{Dragon Ball})                             \\
        \chin{露西} (Lucy)                 & \chin{赛博朋克：边缘行者} (\textit{Cyberpunk: Edgerunners})      & \chin{丁丁} (Tintin)                   & \chin{丁丁历险记} (\textit{The Adventures of Tintin})             \\
        \chin{玛奇玛} (Makima)              & \chin{链锯人} (\textit{Chainsaw Man})                      & \chin{薇尔莉特·伊芙加登} (Violet Evergarden) & \chin{紫罗兰永恒花园} (\textit{Violet Evergarden})                  \\
        \chin{御坂美琴} (Mikoto Misaka)      & \chin{某科学的超电磁炮} (\textit{A Certain Scientific Railgun}) & \chin{维尼} (Winnie the Pooh)          & \chin{小熊维尼} (\textit{Winnie the Pooh})                       \\
        \chin{宫水三叶} (Mitsuha Miyamizu)   & \chin{你的名字} (\textit{Your Name})                        & \chin{金刚狼} (Wolverine)               & \chin{金刚狼系列} (\textit{Wolverine Series})                     \\ \hline
        \end{tabular}%
    }
    \caption{Characters from anime and comics in \textit{RoleEval-Global}.}
    \label{tab:anime_comics_char_list}
\end{table*}

\begin{table*}[]
    \centering
    \resizebox{\textwidth}{!}{%
        \begin{tabular}{cc|cc}
        \hline
        \textbf{Name}                         & \textbf{Source}                                  & \textbf{Name}                       & \textbf{Source}                                     \\ \hline
        \chin{亚伦·霍奇纳} (Aaron Hotchner)        & \chin{犯罪心理} (\textit{Criminal Minds})            & \chin{莱昂} (Léon)                    & \chin{这个杀手不太冷} (\textit{Léon: The Professional})    \\
        \chin{阿蕾莎} (Alessa Gillespie)         & \chin{寂静岭} (\textit{Silent Hill})                & \chin{万磁王} (Magneto)                & \chin{X战警} (\textit{X-Men})                         \\
        \chin{巴斯光年} (Buzz Lightyear)          & \chin{玩具总动员} (\textit{Toy Story})                & \chin{威震天} (Megatron)               & \chin{变形金刚} (\textit{Transformers})                 \\
        \chin{美国队长} (Captain America)         & \chin{复仇者联盟} (\textit{The First Avenger})       & \chin{迈克尔·斯科菲尔德} (Michael Scofield) & \chin{越狱} (\textit{Prison Break})                   \\
        \chin{卡莉·马西森} (Carrie Mathison)       & \chin{国土安全} (\textit{Homeland})                  & \chin{米格尔·里韦拉斯} (Miguel Rivera)     & \chin{寻梦环游记} (\textit{Coco})                        \\
        \chin{克莱尔·邓菲} (Claire Dunphy)         & \chin{摩登家庭} (\textit{Modern Family})             & \chin{三澄美琴} (Mikoto Misumi)         & \chin{非自然死亡} (\textit{Unnatural})                   \\
        \chin{丹妮莉丝·坦格利安} (Daenerys Targaryen) & \chin{权力的游戏} (\textit{Game of Thrones})          & \chin{永尾完治} (Nagao Kanji)           & \chin{东京爱情故事} (\textit{Tokyo Love Story})           \\
        \chin{达里尔·迪克森} (Daryl Dixon)          & \chin{行尸走肉} (\textit{The Walking Dead})          & \chin{尼奥} (Neo)                     & \chin{黑客帝国} (\textit{The Matrix})                   \\
        \chin{戴安娜·普林斯} (Diana Prince)         & \chin{神奇女侠} (\textit{Wonder Woman})              & \chin{艾莎公主} (Princess Elsa)         & \chin{冰雪奇缘} (\textit{Frozen})                       \\
        \chin{伊丽莎白·斯旺} (Elizabeth Swann)      & \chin{加勒比海盗} (\textit{Pirates of the Caribbean}) & \chin{瑞秋·格林} (Rachel Green)         & \chin{老友记} (\textit{Friends})                       \\
        \chin{艾伦·雷普莉} (Ellen Ripley)          & \chin{异形系列} (\textit{Alien series})              & \chin{拉斐尔（忍者神龟）} (Raphael)          & \chin{忍者神龟} (\textit{Teenage Mutant Ninja Turtles}) \\
        \chin{阿甘} (Forrest Gump)              & \chin{阿甘正传} (\textit{Forrest Gump})              & \chin{谢尔顿·库珀} (Sheldon Cooper)      & \chin{生活大爆炸} (\textit{The Big Bang Theory})         \\
        \chin{弗兰西斯·安德伍德} (Francis Underwood)  & \chin{纸牌屋} (\textit{House of Cards})             & \chin{辛巴} (Simba)                   & \chin{狮子王} (\textit{The Lion King})                 \\
        \chin{格鲁} (Gru)                       & \chin{神偷奶爸} (\textit{Despicable Me})             & \chin{托尼·史塔克} (Tony Stark)          & \chin{钢铁侠} (\textit{Iron Man})                      \\
        \chin{哈利·波特} (Harry Potter)           & \chin{哈利波特} (\textit{Harry Potter})              & \chin{迪迦奥特曼} (Ultraman Tiga)        & \chin{迪迦奥特曼} (\textit{Ultraman Tiga})               \\
        \chin{印第安纳·琼斯} (Indiana Jones)        & \chin{夺宝奇兵} (\textit{Indiana Jones})             & \chin{毒液} (Venom)                   & \chin{毒液系列} (\textit{Venom series})                 \\
        \chin{杰克·萨利} (Jake Sully)             & \chin{阿凡达} (\textit{Avatar})                     & \chin{维托·柯里昂} (Vito Corleone)       & \chin{教父} (\textit{The Godfather})                  \\
        \chin{詹姆斯·邦德} (James Bond)            & \chin{007系列} (\textit{James Bond series})        & \chin{瓦力} (WALL-E)                  & \chin{机器人总动员} (\textit{WALL-E})                     \\
        \chin{约翰·麦克莱恩} (John McClane)         & \chin{龙胆虎威} (\textit{Die Hard})                  & \chin{沃尔特·怀特} (Walter White)        & \chin{绝命毒师} (\textit{Breaking Bad})                 \\
        \chin{卢克·天行者} (Luke Skywalker)        & \chin{星球大战} (\textit{Star Wars})                 & \chin{甄}\abchin{嬛} (Zhen Huan)               & \chin{甄}\abchin{嬛}\chin{传} (\textit{Empresses in the Palace})       \\ \hline
        \end{tabular}%
    }
    \caption{Characters from movies and TV series in \textit{RoleEval-Global}.}
    \label{tab:movie_tv_char_list}
\end{table*}

\begin{table*}[]
    \centering
    \resizebox{\textwidth}{!}{%
        \begin{tabular}{cc|cc}
        \hline
        \textbf{Name}                                 & \textbf{Source}                                          & \textbf{Name}                                       & \textbf{Source}                                 \\ \hline
        \chin{亚特鲁·克里斯汀} (Adol Christin)               & \chin{伊苏系列} (\textit{Ys Series})                         & \chin{桐生一马} (Kazuma Kiryu)                          & \chin{如龙系列} (\textit{Yakuza Series})            \\
        \chin{杀手47} (Agent 47)                        & \chin{杀手系列} (\textit{Hitman Series})                     & \chin{奎托斯} (Kratos)                                 & \chin{战神} (\textit{God of War})                 \\
        \chin{艾丹} (Aidan)                             & \chin{暗黑破坏神} (\textit{Diablo})                           & \chin{约翰-117} (Master Chief Petty Officer John-117) & \chin{光环} (\textit{Halo})                       \\
        \chin{艾登·皮尔斯} (Aiden Pearce)                  & \chin{看门狗} (\textit{Watch Dogs})                         & \chin{洛克人} (Mega Man)                               & \chin{洛克人系列} (\textit{Mega Man Series})         \\
        \chin{艾伦·韦克} (Alan Wake)                      & \chin{心灵杀手} (\textit{Alan Wake})                         & \chin{米亚·卡恩斯坦} (Mia Karnstein)                      & \chin{噬血代码} (\textit{Code Vein})                \\
        \chin{埃洛伊} (Aloy)                             & \chin{地平线系列} (\textit{Horizon Series})                   & \chin{蒙葛特} (Morgott)                                & \chin{艾尔登法环} (\textit{Elden Ring})              \\
        \chin{阿尔萨斯·米奈希尔} (Arthas Menethil)            & \chin{魔兽} (\textit{Warcraft})                            & \chin{纳西妲} (Nahida)                                 & \chin{原神} (\textit{Genshin Impact})             \\
        \chin{亚瑟·摩根} (Arthur Morgan)                  & \chin{荒野大镖客} (\textit{Red Dead Redemption})              & \chin{内森·德雷克} (Nathan Drake)                        & \chin{神秘海域系列} (\textit{Uncharted Series})       \\
        \chin{贝优妮塔} (Bayonetta)                       & \chin{猎天使魔女} (\textit{Bayonetta})                        & \chin{山姆·费舍尔} (Sam Fisher)                          & \chin{细胞分裂} (\textit{Splinter Cell})            \\
        \chin{春丽} (Chun-Li)                           & \chin{街头霸王} (\textit{Street Fighter})                    & \chin{萨姆斯·阿兰} (Samus Aran)                          & \chin{银河战士系列} (\textit{Metroid Series})         \\
        \chin{薛帕德} (Commander Shepard)                & \chin{质量效应系列} (\textit{Mass Effect Series})              & \chin{马力欧} (Mario)                       & \chin{超级马力欧兄弟} (\textit{Super Mario Bros}) \\
        \chin{但丁} (Dante)                             & \chin{鬼泣} (\textit{Devil May Cry})                       & \chin{希侬·埃梅利斯} (Shionne Imeris)                     & \chin{破晓传说} (\textit{Tales of Arise})           \\
        \chin{龙裔} (Dragonborn)                        & \chin{上古卷轴5} (\textit{The Elder Scrolls V: Skyrim})      & \chin{索利德·斯内克} (Solid Snake)                        & \chin{合金装备系列} (\textit{Metal Gear Series})      \\
        \chin{艾黛尔贾特·冯·弗雷斯贝尔古} (Edelgard von Hresvelg) & \chin{火焰之纹章：风花雪月} (\textit{Fire Emblem: Three Houses})   & \chin{索拉} (Sora)                                    & \chin{王国之心} (\textit{Kingdom Hearts})           \\
        \chin{爱莉希雅} (Elysia)                          & \chin{崩坏3} (\textit{Honkai Impact 3rd})                  & \chin{猎空} (Tracer)                                  & \chin{守望先锋} (\textit{Overwatch})                \\
        \chin{艾吉奥} (Ezio Auditore da Firenze)         & \chin{刺客信条2} (\textit{Assassin's Creed II})              & \chin{崔佛·菲利普} (Trevor Philips)                      & \chin{GTA5} (\textit{Grand Theft Auto V})       \\
        \chin{杰洛特} (Geralt of Rivia)                  & \chin{巫师} (\textit{The Witcher})                         & \chin{瓦尔基里} (Valkyrie)                              & \chin{Apex英雄} (\textit{Apex Legends})           \\
        \chin{GLaDOS} (GLaDOS)                        & \chin{传送门} (\textit{Portal})                             & \chin{威廉·亚当斯} (William Adams)                       & \chin{仁王系列} (\textit{Nioh Series})              \\
        \chin{苇名一心} (Isshin Ashina)                   & \chin{只狼：影逝二度} (\textit{Sekiro: Shadows Die Twice})      & \chin{尤尔哈2B} (YoRHa No.2 Type B (2B))               & \chin{尼尔：机械纪元} (\textit{NieR: Automata})        \\
        \chin{詹姆斯·波特} (James Porter)                  & \chin{彩虹六号：围攻} (\textit{Tom Clancy's Rainbow Six Siege}) & \chin{塞尔达} (Zelda)                                  & \chin{塞尔达传说} (\textit{The Legend of Zelda})     \\ \hline
        \end{tabular}%
    }
    \caption{Characters from games in \textit{RoleEval-Global}.}
    \label{tab:game_char_list}
\end{table*}

\begin{table*}[]
    \centering
    \resizebox{\textwidth}{!}{%
        \begin{tabular}{cc|cc}
        \hline
            \textbf{Name} & \textbf{Source} & \textbf{Name} & \textbf{Source} \\ \hline
            \chin{爱丽丝} (Alice) & \chin{爱丽丝梦游仙境} (\textit{Alice's Adventures in Wonderland}) & \chin{利奥波德·布卢姆} (Leopold Bloom) & \chin{尤利西斯} (\textit{Ulysses}) \\ 
            \chin{阿迪克斯·芬奇} (Atticus Finch) & \chin{杀死一只知更鸟} (\textit{To Kill a Mockingbird}) & \chin{林黛玉} (Lin Daiyu) & \chin{红楼梦} (\textit{Dream of the Red Chamber}) \\ 
            \chin{曹操} (Cao Cao) & \chin{三国演义} (\textit{Romance of the Three Kingdoms}) & \chin{达西} (Mr. Darcy) & \chin{傲慢与偏见} (\textit{Pride and Prejudice}) \\ 
            \chin{亚哈船长} (Captain Ahab) & \chin{白鲸} (\textit{Moby-Dick}) & \chin{奥利弗·崔斯特} (Oliver Twist) & \chin{雾都孤儿} (\textit{Oliver Twist}) \\ 
            \chin{尼摩船长} (Captain Nemo) & \chin{海底两万里} (\textit{Twenty Thousand Leagues Under the Seas}) & \chin{保罗·厄崔迪} (Paul Atreides) & \chin{沙丘} (\textit{Dune}) \\ 
            \chin{爱德蒙·邓蒂斯} (Edmond Dantès) & \chin{基督山伯爵} (\textit{The Count of Monte Cristo}) & \chin{保尔·柯察金} (Paul Korchagin) & \chin{钢铁是怎样炼成的} (\textit{How the Steel Was Tempered}) \\ 
            \chin{爱德华·卡伦} (Edward Cullen) & \chin{暮光之城} (\textit{Twilight}) & \chin{波西·杰克逊} (Percy Jackson) & \chin{波西杰克逊系列} (\textit{Percy Jackson \& the Olympians}) \\ 
            \chin{葛朗台} (Eugénie Grandet) & \chin{欧也妮·葛朗台} (\textit{Eugénie Grandet}) & \chin{皮埃尔·别祖霍夫} (Pierre Bezukhov) & \chin{战争与和平} (\textit{War and Peace}) \\ 
            \chin{芬恩·麦克库尔} (Finn McCool) & \chin{芬尼亚传奇} (\textit{The Fenian Cycle}) & \chin{卡西莫多} (Quasimodo) & \chin{巴黎圣母院} (\textit{The Hunchback of Notre-Dame}) \\ 
            \chin{甘道夫} (Gandalf) & \chin{霍比特人} (\textit{The Hobbit}) & \chin{鲁滨逊·克鲁索} (Robinson Kreutznaer) & \chin{鲁宾逊漂流记} (\textit{Robinson Crusoe}) \\ 
            \chin{格列佛} (Gulliver) & \chin{格列佛游记} (\textit{Gulliver's Travels}) & \chin{桐原亮司} (Ryōji Kirihara) & \chin{白夜行} (\textit{Journey Under the Midnight Sun}) \\ 
            \chin{希斯克利夫} (Heathcliff) & \chin{呼啸山庄} (\textit{Wuthering Heights}) & \chin{索伦} (Sauron) & \chin{魔戒} (\textit{The Lord of the Rings}) \\ 
            \chin{赫尔克里·波洛} (Hercule Poirot) & \chin{东方列车谋杀案} (\textit{Murder on the Orient Express}) & \chin{斯嘉丽·奥哈拉} (Scarlett O'Hara) & \chin{飘} (\textit{Gone with the Wind}) \\ 
            \chin{霍尔顿·考尔菲德} (Holden Caulfield) & \chin{麦田里的守望者} (\textit{The Catcher in the Rye}) & \chin{夏洛克·福尔摩斯} (Sherlock Holmes) & \chin{福尔摩斯探案集} (\textit{The Adventures of Sherlock Holmes}) \\ 
            \chin{简·爱} (Jane Eyre) & \chin{简·爱} (\textit{Jane Eyre}) & \chin{宋江} (Song Jiang) & \chin{水浒传} (\textit{Water Margin}) \\ 
            \chin{盖茨比} (Jay Gatsby) & \chin{了不起的盖茨比} (\textit{The Great Gatsby}) & \chin{孙悟空} (Sun Wukong) & \chin{西游记} (\textit{Journey to the West}) \\ 
            \chin{冉·阿让} (Jean Valjean) & \chin{悲惨世界} (\textit{Les Misérables}) & \chin{汤姆·索亚} (Tom Sawyer) & \chin{汤姆索亚历险记} (\textit{The Adventures of Tom Sawyer}) \\ 
            \chin{于连} (Julien Sorel) & \chin{红与黑} (\textit{The Red and the Black}) & \chin{维克多·弗兰肯斯坦} (Victor Frankenstein) & \chin{弗兰肯斯坦} (\textit{Frankenstein}) \\ 
            \chin{凯特尼斯·伊夫狄恩} (Katniss Everdeen) & \chin{饥饿游戏} (\textit{The Hunger Games}) & \chin{温斯顿·史密斯} (Winston Smith) & \chin{1984} (\textit{1984}) \\ 
            \chin{加贺恭一郎} (Kyōichirō Kaga) & \chin{加贺恭一郎系列} (\textit{Police Detective Kaga series}) & \chin{叶文洁} (Ye Wenjie) & \chin{三体} (\textit{The Three-Body Problem}) \\ \hline
        \end{tabular}
    }
    \caption{Characters from fiction in \textit{RoleEval-Global}.}
    \label{tab:fiction_char_list}
\end{table*}

\begin{table*}[]
    \centering
    \resizebox{\textwidth}{!}{%
        \begin{tabular}{ccc|ccc}
        \hline
        \textbf{Name}                         & \textbf{Source}                                                          & \textbf{Category} & \textbf{Name}                      & \textbf{Source}                                                 & \textbf{Category}    \\ \hline
        \chin{何炅} (He Jiong)                  & -                                                                        & celebrities       & \chin{路明非} (Lu Mingfei)            & \chin{龙族} (\textit{Dragon Raja})                                & fiction              \\
        \chin{周杰伦} (Jay Chou)                 & -                                                                        & celebrities       & \chin{任盈盈} (Ren Yingying)          & \chin{笑傲江湖} (\textit{The Smiling, Proud Wanderer})              & fiction              \\
        \chin{李敏镐} (Lee Min-ho)               & -                                                                        & celebrities       & \chin{颂莲} (Song Lian)              & \chin{妻妾成群} (\textit{Wives and Concubines})                     & fiction              \\
        \chin{雷军} (Lei Jun)                   & -                                                                        & celebrities       & \chin{孙少平} (Sun Shaoping)          & \chin{平凡的世界} (\textit{Ordinary World})                          & fiction              \\
        \chin{李白} (Li Bai)                    & -                                                                        & celebrities       & \chin{唐三} (Tang San)               & \chin{斗罗大陆} (\textit{Douluo Dalu})                              & fiction              \\
        \chin{李荣浩} (Li Ronghao)               & -                                                                        & celebrities       & \chin{祥子} (Xiangzi)                & \chin{骆驼祥子} (\textit{Rickshaw Boy})                             & fiction              \\
        \chin{李宇春} (Li Yuchun)                & -                                                                        & celebrities       & \chin{徐凤年} (Xu Fengnian)           & \chin{雪中悍刀行} (\textit{Sword Snow Stride})                       & fiction              \\
        \chin{李子柒} (Li Ziqi)                  & -                                                                        & celebrities       & \chin{张起灵} (Zhang Qiling)          & \chin{盗墓笔记} (\textit{The Lost Tomb})                            & fiction              \\
        \chin{刘翔} (Liu Xiang)                 & -                                                                        & celebrities       & \chin{张无忌} (Zhang Wuji)            & \chin{倚天屠龙记} (\textit{The Heaven Sword and Dragon Saber})       & fiction              \\
        \chin{刘亦菲} (Liu Yifei)                & -                                                                        & celebrities       & \chin{张小凡} (Zhang Xiaofan)         & \chin{诛仙} (\textit{Zhu Xian})                                   & fiction              \\
        \chin{马伊琍} (Ma Yili)                  & -                                                                        & celebrities       & \chin{奥古斯特·奥特姆} (Augustus Autumn)  & \chin{辐射3} (\textit{Fallout 3})                                 & games                \\
        \chin{秦始皇} (Qin Shi Huang)            & -                                                                        & celebrities       & \chin{克劳德} (Cloud Strife)          & \chin{最终幻想7} (\textit{Final Fantasy VII})                       & games                \\
        \chin{孙燕姿} (Stefanie Sun)             & -                                                                        & celebrities       & \chin{爱梅斯} (Emes)                  & \chin{公主连结Re:Dive} (\textit{Princess Connect! Re:Dive})         & games                \\
        \chin{木村拓哉} (Takuya Kimura)           & -                                                                        & celebrities       & \chin{艾丝蒂尔·布莱特} (Estelle Bright)   & \chin{英雄传说·轨迹系列} (\textit{The Legend of Heroes: Trails Series}) & games                \\
        \chin{王鹤棣} (Wang Hedi)                & -                                                                        & celebrities       & \chin{能天使} (Exusiai)               & \chin{明日方舟} (\textit{Arknights})                                & games                \\
        \chin{王俊凯} (Wang Junkai)              & -                                                                        & celebrities       & \chin{盖伦} (Garen)                  & \chin{英雄联盟} (\textit{League of Legends})                        & games                \\
        \chin{杨振宁} (Yang Chen-Ning)           & -                                                                        & celebrities       & \chin{季沧海} (Ji Canghai)            & \chin{永劫无间} (\textit{Naraka: Bladepoint})                       & games                \\
        \chin{岳云鹏} (Yue Yunpeng)              & -                                                                        & celebrities       & \chin{约翰·普莱斯} (John Price)         & \chin{使命召唤} (\textit{Call of Duty})                             & games                \\
        \chin{张艺谋} (Zhang Yimou)              & -                                                                        & celebrities       & \chin{三岛一八} (Kazuya Mishima)       & \chin{铁拳系列} (\textit{Tekken series})                            & games                \\
        \chin{张艺兴} (Zhang Yixing)             & -                                                                        & celebrities       & \chin{李逍遥} (Li Xiaoyao)            & \chin{仙剑奇侠传} (\textit{Chinese Paladin})                         & games                \\
        \chin{阿尼亚·福杰} (Anya Forger)           & \chin{间谍过家家} (\textit{Spy x Family})                                     & anime and comics  & \chin{马可斯·菲尼克斯} (Marcus Fenix)     & \chin{战争机器系列} (\textit{Gears of War series})                    & games                \\
        \chin{小智} (Ash Ketchum)               & \chin{宝可梦} (\textit{Pokémon})                                            & anime and comics  & \chin{米拉波雷亚斯} (Milla Borealis)     & \chin{怪物猎人} (\textit{Monster Hunter})                           & games                \\
        \chin{荆天明} (Jing Tianming)            & \chin{秦时明月} (\textit{Qin's Moon})                                        & anime and comics  & \chin{蕾米莉亚·斯卡蕾特} (Remilia Scarlet) & \chin{东方project} (\textit{Touhou Project})                      & games                \\
        \chin{四宫辉夜} (Kaguya Shinomiya)        & \chin{辉夜大小姐想让我告白} (\textit{Kaguya-sama: Love Is War})                    & anime and comics  & \chin{赛丽亚·克鲁敏} (Seria Kirmin)      & \chin{地下城与勇士} (\textit{Dungeon Fighter Online})                 & games                \\
        \chin{奇犽·揍敌客} (Killua Zoldyck)        & \chin{全职猎人} (\textit{Hunter × Hunter})                                   & anime and comics  & \chin{史特列洛克} (Strelok)             & \chin{潜行者系列} (\textit{S.T.A.L.K.E.R. series})                   & games                \\
        \chin{马克·霍夫曼} (Marc)                  & \chin{灵笼} (\textit{Ling Cage: Incarnation})                              & anime and comics  & \chin{塔里昂} (Talion)                & \chin{中土世界：战争之影} (\textit{Middle-earth: Shadow of War})         & games                \\
        \chin{中野三玖} (Miku Nakano)             & \chin{五等分的新娘} (\textit{5-toubun no Hanayome})                            & anime and comics  & \chin{狸克} (Tom Nook)               & \chin{动物森友会} (\textit{Animal Crossing})                         & games                \\
        \chin{早濑未沙} (Misa Hayase)             & \chin{超时空要塞} (\textit{Super Dimension Fortress Macross})                 & anime and comics  & \chin{谢衣} (Xie Yi)                 & \chin{古剑奇谭2} (\textit{Gujian2})                                 & games                \\
        \chin{桃园奈奈生} (Nanami Momozono)        & \chin{元气少女缘结神} (\textit{Kamisama Kiss})                                  & anime and comics  & \chin{杨宁} (Yang Ning)              & \chin{剑侠情缘网络版叁} (\textit{Swordsman Online})                     & games                \\
        \chin{喜羊羊} (Pleasant Goat)            & \chin{喜羊羊与灰太狼} (\textit{Pleasant Goat and Big Big Wolf})                 & anime and comics  & \chin{尤里} (Yuri)                   & \chin{红色警戒} (\textit{Command \& Conquer: Red Alert})            & games                \\
        \chin{红细胞} (Red Blood Cell)           & \chin{工作细胞} (\textit{Cells at Work!})                                    & anime and comics  & \chin{都敏俊} (Do Min-joon)           & \chin{来自星星的你} (\textit{My Love from the Star})                  & movies and TV series \\
        \chin{成步堂龙一} (Ryūichi Naruhodō)       & \chin{逆转裁判} (\textit{Ace Attorney})                                      & anime and comics  & \chin{樊胜美} (Fan Shengmei)          & \chin{欢乐颂} (\textit{Ode to Joy})                                & movies and TV series \\
        \chin{埼玉} (Saitama)                   & \chin{一拳超人} (\textit{One Punch Man})                                     & anime and comics  & \chin{东方青苍} (Dongfang Qingcang)               & \chin{苍兰诀} (\textit{Love Between Fairy and Devil})                               & movies and TV series \\
        \chin{菜月昴} (Subaru Natsuki)           & \chin{Re:从零开始的异世界生活} (\textit{Re:Zero - Starting Life in Another World}) & anime and comics  & \chin{范闲} (Fan Xian)               & \chin{庆余年} (\textit{Joy of Life})                               & movies and TV series \\
        \chin{灶门炭治郎} (Tanjiro Kamado)         & \chin{鬼灭之刃} (\textit{Demon Slayer: Kimetsu no Yaiba})                    & anime and comics  & \chin{顾千帆} (Gu Qianfan)            & \chin{梦华录} (\textit{A Dream of Splendor})                       & movies and TV series \\
        \chin{安艺伦也} (Tomoya Aki)              & \chin{路人女主的养成方法} (\textit{Saekano: How to Raise a Boring Girlfriend})    & anime and comics  & \chin{好彩妹} (Hao Cai Mei)           & \chin{搜神传} (\textit{Legend of the Demigods})                    & movies and TV series \\
        \chin{尤尼} (Tsunayoshi ``Tsuna'' Sawada) & \chin{家庭教师} (\textit{Katekyo Hitman Reborn!})                            & anime and comics  & \chin{洪世贤} (Hong Shixian)          & \chin{回家的诱惑} (\textit{The Temptation to Go Home})               & movies and TV series \\
        \chin{涂山苏苏} (Tushan Susu)             & \chin{狐妖小红娘} (\textit{Fox Spirit Matchmaker})                            & anime and comics  & \chin{胡一菲} (Hu Yifei)              & \chin{爱情公寓} (\textit{iPartment})                                & movies and TV series \\
        \chin{月野兔} (Usagi Tsukino)            & \chin{美少女战士} (\textit{Sailor Moon})                                      & anime and comics  & \chin{花千骨} (Hua Qiangu)            & \chin{花千骨} (\textit{The Journey of Flower})                     & movies and TV series \\
        \chin{袁天罡} (Yuan Tiangang)            & \chin{画江湖之不良人} (\textit{Hua Jiang Hu Zhi Bu Liang Ren})                  & anime and comics  & \chin{高进} (Ko Chun)                & \chin{赌神系列} (\textit{God of Gamblers series})                   & movies and TV series \\
        \chin{白素贞} (Bai Suzhen)               & \chin{白蛇传} (\textit{Legend of the White Snake})                          & fiction           & \chin{洛晴川} (Luo Qingchuan)         & \chin{宫锁心玉} (\textit{Palace})                                   & movies and TV series \\
        \chin{陈平安} (Chen Ping'an)             & \chin{剑来} (\textit{Sword of Coming})                                     & fiction           & \chin{半泽直树} (Naoki Hanzawa)        & \chin{半泽直树} (\textit{Naoki Hanzawa})                            & movies and TV series \\
        \chin{楚留香} (Chu Liuxiang)             & \chin{楚留香传奇} (\textit{The Legend of Chu Liuxiang})                       & fiction           & \chin{盛明兰} (Sheng Minglan)         & \chin{知否知否应是绿肥红瘦} (\textit{The Story of MingLan})               & movies and TV series \\
        \chin{崔莺莺} (Cui Yingying)             & \chin{崔莺莺待月西厢记} (\textit{Romance of the Western Chamber})                & fiction           & \chin{武媚娘} (Wu Meiniang)           & \chin{武媚娘传奇} (\textit{The Empress of China})                    & movies and TV series \\
        \chin{戴凤莲} (Dai Fenglian)             & \chin{红高粱家族} (\textit{Red Sorghum Family})                               & fiction           & \chin{小燕子} (Xiao Yanzi)            & \chin{还珠格格} (\textit{My Fair Princess})                         & movies and TV series \\
        \chin{封三娘} (Feng Sanniang)            & \chin{聊斋志异} (\textit{Strange Stories from a Chinese Studio})             & fiction           & \chin{姚金铃} (Yao Jinling)           & \chin{宫心计} (\textit{Beyond the Realm of Conscience})            & movies and TV series \\
        \chin{何以琛} (He Yichen)                & \chin{何以笙箫默} (\textit{Silent Separation})                                & fiction           & \chin{余则成} (Yu Zecheng)            & \chin{潜伏} (\textit{Lurk})                                       & movies and TV series \\
        \chin{胡八一} (Hu Bayi)                  & \chin{鬼吹灯} (\textit{Ghost Blows Out the Light})                          & fiction           & \chin{余罪} (Yu Zui)                 & \chin{余罪} (\textit{Yu Zui})                                                & movies and TV series \\
        \chin{花城} (Hua Cheng)                 & \chin{天官赐福} (\textit{Heaven Official's Blessing})                        & fiction           & \chin{张麻子} (Zhang Mazi)            & \chin{让子弹飞} (\textit{Let the Bullets Fly})                      & movies and TV series \\
        \chin{孔乙己} (Kong Yiji)                & \chin{孔乙己} (\textit{Kong Yiji})                                          & fiction           & \chin{紫霞仙子} (Zixia Fairy)          & \chin{大话西游} (\textit{A Chinese Odyssey})                        & movies and TV series \\ \hline
        \end{tabular}%
    }
    \caption{Characters in \textit{RoleEval-Chinese}.}
    \label{tab:RoleEval_chinese_list}
\end{table*}
\end{document}